# Abnormal Head Movements in Neurological Conditions: A Knowledge-Based Dataset with Application to Cervical Dystonia

Saja Al-Dabet*, Sherzod Turaev, Nazar Zaki

College of Information Technology, United Arab Emirates University, Al Ain, United Arab Emirates

700039885@uaeu.ac.ae, sherzod@uaeu.ac.ae , nzaki@uaeu.ac.ae

## Abstract

Abnormal head movements (AHMs) manifest across a broad spectrum of neurological disorders, however the absence of a multi-condition resource integrating kinematic measurements, clinical severity scores, and patient demographics constitutes a persistent barrier to the development of AI-driven diagnostic tools. To address this gap, this study introduces NeuroPose-AHM, a knowledge-based dataset of neurologically induced AHMs constructed through a multi-LLM extraction framework applied to 1430 peer-reviewed publications. The dataset contains 2756 patient-group-level records spanning 57 neurological conditions, derived from 846 AHM-relevant papers. Inter-LLM reliability analysis confirms robust extraction performance, with study-level classification achieving strong agreement (κ = 0.822). To demonstrate the dataset's analytical utility, a four-task framework is applied to cervical dystonia (CD), the condition most directly defined by pathological head movement. First, Task 1 performs multi-label AHM type classification (F1 = 0.856). Task 2 constructs the Head–Neck Severity Index (HNSI), a unified metric that normalizes heterogeneous clinical rating scales. The clinical relevance of this index is then evaluated in Task 3, where HNSI is validated against real-world CD patient data, with aligned severe-band proportions (6.7%) providing a preliminary plausibility indication for index calibration within the high severity range. Finally, Task 4 performs bridge analysis between movement-type probabilities and HNSI scores, producing significant correlations ($p < 0.001$). These results demonstrate the analytical utility of NeuroPose-AHM as a structured, knowledge-based resource for neurological AHM research. The NeuroPose-AHM dataset is publicly available on Zenodo (https://doi.org/10.5281/zenodo.19386862).

## 1. Introduction

Body language is a fundamental channel of nonverbal clinical communication in which gestures, posture, and movement convey information about an individual's physical and psychological state [1]. Among these, abnormal head position (AHPOS) — characterized by a deviation of the head from its neutral alignment — represents a clinically significant sign observed across a broad spectrum of neurological, ocular, and musculoskeletal conditions [2,3]. AHPOS encompasses two distinct manifestations: abnormal head postures (AHP), which are sustained positional deviations, and abnormal head movements (AHM), which involve involuntary, repetitive, or episodic motions such as tremor, nodding, and shaking, frequently arising from neurological dysfunction [4].

Neurological conditions constitute a significant etiology of AHPOS, arising from disruptions within the motor control pathways responsible for regulating head and neck position. Cervical Dystonia (CD) is a leading neurological cause of AHPOS. It is characterized by involuntary contractions of the neck muscles that produce abnormal head postures and repetitive movements, which typically worsen during voluntary activity [5]. Its primary and direct manifestation as pathological AHM distinguishes it from conditions in which head involvement is secondary or incidental [6,7]. This specificity makes CD one of the most suitable conditions for developing automated diagnostic and severity quantification frameworks within the broader AHP analysis context [8].

These motor disturbances originate from impaired signaling within the basal ganglia-thalamo-cortical circuit, a pathway central to the regulation of voluntary movement [9,10]. CD patients exhibit a characteristic co-occurrence of multiple movement patterns including rotation (torticollis), lateral tilt (laterocollis), anterior flexion (anterocollis), and posterior extension (retrocollis) [7], as illustrated in Figure 1. Other conditions such as Parkinson's disease, Huntington's disease, and multiple sclerosis also present with AHM as a primary or secondary manifestation [11–13].

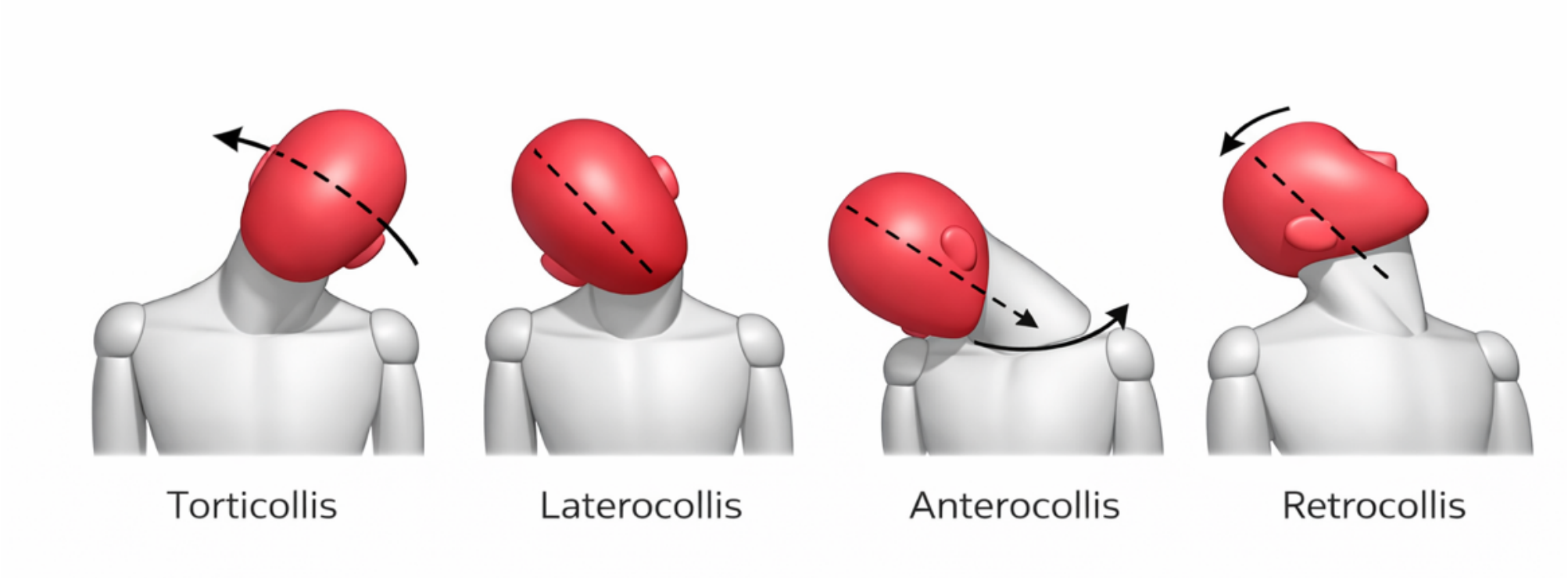

*Figure 1: Primary AHPOS Manifestations in Cervical Dystonia*

Despite advances in AI and deep learning, diagnostic tools for AHPOS remain limited in scope. A primary constraint is the absence of comprehensive datasets encompassing the full diversity of AHM manifestations across neurological conditions. Conditions associated with AHPOS have been examined independently across the literature, with no systematic effort to consolidate their clinical knowledge into a unified resource. This challenge is compounded by widespread inconsistency in clinical severity measurement, with conditions documented using heterogeneous rating instruments including Toronto Western Spasmodic Torticollis Rating Scale (TWSTRS), the Burke-Fahn-Marsden Dystonia Rating Scale (BFMDRS), the Tsui Scale, and the Global Dystonia Rating Scale (GDRS), each employing distinct scoring ranges that render direct cross-study comparison invalid without scale-aware normalization. Movement patterns are further documented using inconsistent terminology across studies, with no unified representational standard enabling systematic cross-condition comparison.

In this paper, we present NeuroPose-AHM, a structured, literature-derived, multi-condition dataset of neurologically induced AHMs, constructed through a multi-LLM extraction framework applied to 1,430 peer-reviewed publications. To demonstrate its analytical utility, a four-task framework is applied to cervical dystonia (CD), the condition most directly defined by pathological head movement, integrating multi-label AHM classification, unified severity indexing through the Head–Neck Severity Index (HNSI), clinical validation against real-world patient data, and bridge analysis cross-validating movement and severity outputs. The contributions of this work can be summarized as follows:

1. NeuroPose-AHM, a knowledge-based dataset of neurologically induced AHMs constructed through a multi-LLM extraction framework. The dataset captures quantitative kinematic measurements, clinical severity scores, and patient demographics across 57 neurological conditions.
2. The Head–Neck Severity Index (HNSI), a normalization metric that standardizes heterogeneous clinical rating scales into a comparable severity score, enabling cross-study severity analysis independent of the original assessment instrument.
3. A four-task analytical framework applied to cervical dystonia, integrating multi-label AHM classification, HNSI-based severity indexing, clinical validation using real-world patient data, and bridge analysis, demonstrating the dataset's analytical utility.

The remainder of this paper is organized as follows. Section 2 reviews the literature on neurological conditions associated with AHMs and existing AI approaches. Section 3 describes the methodology, including the NeuroPose-AHM dataset construction pipeline and the four-task framework applied to cervical dystonia. Section 4 presents the results and discussion. Section 5 concludes the paper and outlines future work.

## 2. Literature Review

The neurological conditions associated with abnormal head movements have been the subject of extensive survey and systematic review literature. This section reviews the key findings relevant to the major conditions in this domain, highlighting the gaps that motivate the construction of NeuroPose-AHM.

### 2.1 Neurological Conditions Associated with Abnormal Head Movements: Clinical Overview

Among neurological conditions associated with AHM, CD and essential tremor (ET) are the most prominent. Defazio et al. (2013) reported prevalence estimates ranging from 20 to 4,100 per million across 16 epidemiological studies, with variability largely attributed to methodological differences [14]. Tiderington et al. (2013) demonstrated that despite its identifiable clinical features, patients saw a mean of 3.5 providers over 44 months before receiving a confirmed diagnosis, reflecting both phenotypic complexity and the absence of objective biomarkers [15]. The MDS Dystonia Study Group subsequently formalized diagnostic criteria and classification for isolated CD, however acknowledged that diagnosis remains dependent on clinical expertise and is not consistently reproducible across centers [16].

ET is also among the most common movement disorders globally. Song et al. (2021) performed a systematic review of 29 studies, reporting a pooled prevalence of 0.32% across all ages, rising to 2.87% in individuals aged 80 years and older. Head tremor has been reported in approximately 50% of ET cases [17]. A key diagnostic challenge is the frequent confusion between ET and Parkinson's disease; Jankovic et al. (2006) found that 37% of patients with a prior ET diagnosis were in fact misdiagnosed, with Parkinson's disease being the most common true diagnosis [18]. Louis et al. (2025) subsequently confirmed this overlap in a systematic review of 33 studies, reporting an elevated risk of conversion from ET to Parkinson's disease of approximately 4–5 times the baseline rate [19].

The application of AI to neurological conditions associated with AHMs has been examined across several multi-condition systematic reviews and studies. Vizcarra et al. (2024) conducted a systematic review of AI in hyperkinetic movement disorders including dystonia, essential tremor, and Parkinson's disease across 55 studies involving 11,946 subjects, finding that datasets were predominantly small, single-center, and single-condition, limiting model generalizability [20]. Similarly, a systematic review of video-based data-driven models covering tremor, dystonia, and Parkinson's disease concluded that most available modalities remain focused on Parkinson's disease and are largely unexplored for other movement disorders [21].

## 2.2 AI-Based Detection of Cervical Dystonia

Early AI-based approaches to CD detection leveraged neuroimaging. Valeriani and Simonyan (2020) developed DystoniaNet, a DL platform trained on raw structural MRI from 612 subjects, identifying a microstructural neural network biomarker and achieving 100% diagnostic accuracy for CD and 98.8% overall across three focal dystonia subtypes [22]. Attention subsequently shifted to contactless, video- and depth-sensor-based quantification of head motor symptoms. Nakamura et al. (2019) demonstrated early feasibility using a Kinect depth sensor to semi-automatically compute TWSTRS severity scores from head angles in 30 CD patients, reporting a significant correlation with neurologist-assigned ratings (r = 0.596) [23]. Ye et al. (2022) extended this direction using Azure Kinect RGB-D cameras to automatically score TWSTRS subscales, though validation remained limited to eight participants [24].

Zhang et al. (2022) introduced the Computational Motor Objective Rater (CMOR), a computer vision and machine learning system that extracted 3D head pose angles from conventional video recordings across 185 CD patients at 10 sites. Their approach demonstrated strong convergent validity with TWSTRS clinical severity ratings (rho = 0.59–0.68) [25]. The same framework was extended to head tremor quantification in 93 CD patients, yielding a correlation of rho = 0.54 with clinical ratings and identifying pitch as the dominant tremor axis in 50% of cases [26]. Peach et al. (2024) further advanced this direction with a visual-perceptive deep learning framework applied across seven academic centers. Their approach extracted kinematic pathosignatures from standard clinical video that encoded disease severity, dystonia subtype, and response to deep brain stimulation [27]. A recent systematic review of computer vision technologies in movement disorders confirmed that CD-specific approaches show strong alignment between automated measurements and clinical assessments, but highlighted persistent heterogeneity in video settings and the absence of standardized recording guidelines across studies [28].

Across both clinical and AI-based literature, several gaps constrain progress in neurological AHM analysis. First, existing AI approaches treat neurological conditions in isolation and rely on direct patient data acquisition through video, sensors, or neuroimaging, lacking a unified resource that integrates evidence across neurological conditions. Second, quantitative AHM characterization—including kinematic features such as frequency, velocity, and amplitude—remains sparsely studied, and no multi-condition, literature-derived resource consolidating these measurements with clinical severity assessments. Third, clinical severity evaluation remains distributed across heterogeneous rating scales, preventing cross-study comparison independent of the original assessment instrument. These gaps hinder the development of generalizable AI-driven diagnostic frameworks for neurological AHM analysis.

## 3. Proposed Methodology

### *3.1 NeuroPose-AHM: Knowledge-Based Neurological AHM Dataset Generation*

Despite extensive research on neurological conditions that cause AHM, there remains a lack of comprehensive datasets integrating quantitative kinematic measurements, clinical assessments, and patient-level characteristics required for the development of AI-based diagnostic frameworks. NeuroPose-AHM addresses specifically the neurological subset of AHM, focusing on AHM as the primary dynamic manifestation of the included conditions. The dataset is constructed as an expert-level, knowledge-based resource derived from the neurological research literature. It aims to provide a structured and analytical foundation for the automated diagnosis of neurological disorders underlying AHM. Dataset generation is conducted through a multi-stage extraction framework that integrates multi LLM-based information extraction with iterative refinement steps, applied to 1430 publications identified through a systematic literature review. This approach replicates expert-level clinical judgment at scale, addressing the challenges of structured data extraction from unstructured medical texts while ensuring reliability in capturing the clinical knowledge embedded within the literature.

### *3.1.1 Literature Search Strategy*

An initial literature search is conducted using research databases, including PubMed, Scopus, Google Scholar, and Web of Science. The objective is to identify neurological conditions that may influence patients' movements, particularly involving head movements. A combination of condition names and terms related to AHM or involuntary movements is utilized. Boolean operators such as "AND" and "OR" are employed to refine the queries, for example, "Parkinson's Disease AND anterocollis OR torticollis" and "Multiple Sclerosis AND head tremor". A total of 835 medical research papers is retrieved during this process.

Subsequently, a parameter-focused literature retrieval is applied to specifically identify studies reporting quantitative AHM characteristics. A systematic PubMed search is conducted using an expanded disease and disorder query list in combination with automated title and abstract filters. Head-related terms, specifically "head," "neck," or "cervical," are used as mandatory keywords to ensure relevance. Four structured parameter categories are targeted: frequency, velocity, amplitude, latency, and an "All Parameters" group requiring co-occurrence of all four measurement types in the abstract. Approximately 595 additional papers are collected during this stage. After duplicate records are removed, a total of 1430 papers is retained across both stages. Following post-extraction quality filtering, 1,200 distinct studies are confirmed, as summarized in Table 1.

### *3.1.2 Study Selection and Classification Criteria*

The collected papers are categorized according to neurological disease or disorder classification. This distinction is based on differences in etiology, structural pathology, and functional impact. Neurological diseases typically involve identifiable structural or pathological changes within the nervous system and often have a well-defined cause, such as Alzheimer's or Parkinson's disease. In contrast, neurological disorders refer to disturbances in nervous system function that may occur without consistent structural abnormalities, often reflecting dysfunction in neural circuits or developmental processes, as seen in dystonia and essential tremor [29,30]. While all diseases involve disordered function, not all disorders meet the criteria for disease, as many are defined primarily by clinical symptoms rather than confirmed biological pathology.

Table 1 summarizes the number of distinct papers identified in each category, along with the total number of extracted rows. Across all collected papers, 1200 unique studies are confirmed following post-extraction quality filtering. When focusing specifically on AHM, a total of 846 distinct papers is identified, representing a substantial subset of the overall dataset. Collectively, the NeuroPose-AHM dataset comprises 2,756 patient-group-level records spanning 57 unique neurological conditions across both disease and disorder categories.

*Table 1: Distinct papers and total extracted rows across categories.*

| Category | Distinct Papers | Total Rows |
|---|---|---|
| All Papers | 1200 | 4446 |
| Disorder | 709 | 2636 |
| Disease | 414 | 1142 |
| Disorder with AHM | 650 | 2272 |
| Disease with AHM | 196 | 491 |
| All Papers with AHM | 846 | 2756 |

### *3.1.3 Document Preprocessing and Standardization*

Prior to structured extraction, all collected research papers are converted to markdown format using Claude Sonnet 4.5 [31] to ensure consistent text representation across diverse publication sources. This conversion preserves key paper components, including section headings, subsections, tables, and paragraph boundaries, while removing formatting artifacts that may interfere with language model processing. Figure descriptions are also generated and incorporated into the markdown text, as figures often contain clinically relevant information such as patient images showing abnormalities, kinematic measurement plots, and diagnostic imaging results related to abnormal head movement characterization. Each paper is assigned a unique identifier corresponding to its classification within the disease or disorder taxonomy established during the study selection phase.

### *3.1.4 Prompt Engineering and Schema Design*

A comprehensive prompt template is developed to guide the extraction process, explicitly defining the target JSON schema, extraction rules, and clinical terminology conventions relevant to AHM in neurological conditions. The extraction generates a dataset capturing quantitative kinematic measurements, clinical scale scores, movement patterns, and patient demographics across neurological conditions. This dataset forms the foundation for developing diagnostic frameworks for AHM related to neurological conditions. The schema is structured hierarchically to capture multiple levels of clinical information, as detailed in Table 2.

*Table 2: JSON Schema Structure for Clinical Data Extraction*

| Schema Level | Key Fields |
|---|---|
| Study-Level Metadata | paper id, study title, study type, total sample size, study age range, study gender distribution |
| Patient Group Level | group id, condition name, condition category, n patients, age, age range, gender, gender distribution, causes ahm, head symptoms, general symptoms |
| Head Movement Characterization | type, direction, laterality, degree, frequency, consistency, pattern |
| Quantitative Measurements | measurement performed, measurement system, measurement location, frequency value, frequency unit, velocity value, velocity unit, amplitude value, amplitude unit, amplitude direction, latency value, latency unit |
| Pain Assessment | pain present, pain severity, pain severity scale, pain location, pain characteristics |
| Eye | eye abnormalities |
| Clinical Scale Assessments | scale name, scale type, subscale, score range, baseline value, post treatment value, change value, p value, measurement timepoint |

The hierarchical organization of the schema begins at the study level, capturing publication identifiers, study design classification, and aggregate demographic characteristics. At the patient group level, separate entries are created for each distinct group or individual patient to prevent inappropriate data aggregation that could obscure clinically meaningful variations in phenotypes, severity, or treatment responses. Building on this structure, each patient group entry documents AHM characteristics by specifying movement type, and is complemented by quantitative measurement fields for frequency, velocity, amplitude, and latency, along with the measurement technology employed. In addition, pain assessment fields record the presence of pain and severity ratings, while eye abnormality fields capture ocular anomalies associated with head movement. Finally, clinical scale assessment fields are included to accommodate neurological outcome measures, with subscale specification and temporal tracking of both baseline and post-treatment values.

To ensure reliable extraction across all schema levels, the prompt includes specific instructions to maintain consistency and accuracy. LLMs are instructed to extract only information explicitly stated in the text rather than making inferences. To preserve numerical precision, numbers are copied exactly as written without rounding or converting units. Regarding data structure, an important rule requires creating separate entries for each distinct patient or group to avoid losing clinically meaningful variation. This schema and prompt enable robust extraction and analysis of AHM patterns across neurological conditions.

#### *3.1.5 Multi-Model Extraction Framework with Iterative Refinement*

To enable robust information extraction from clinical literature, a multi-model framework is implemented involving two LLMs: GPT 5.1 [32,33] and Claude Sonnet 4.5. Figure 2 presents an overview of the proposed multi-model extraction framework with iterative refinement. Both models are independently deployed on each paper to extract structured information according to the predefined schema and prompt. This approach is designed to mitigate individual model biases and to leverage complementary strengths across distinct architectures, thereby improving the reliability and comprehensiveness of the resulting clinical dataset. Each model receives the same prompt, which includes standardized extraction instructions and the markdown-formatted paper text. Temperature parameters are set to 0.0 to eliminate stochastic variability and ensure consistent, reproducible outputs. The extraction process targets all schema-defined information categories across all collected research papers.

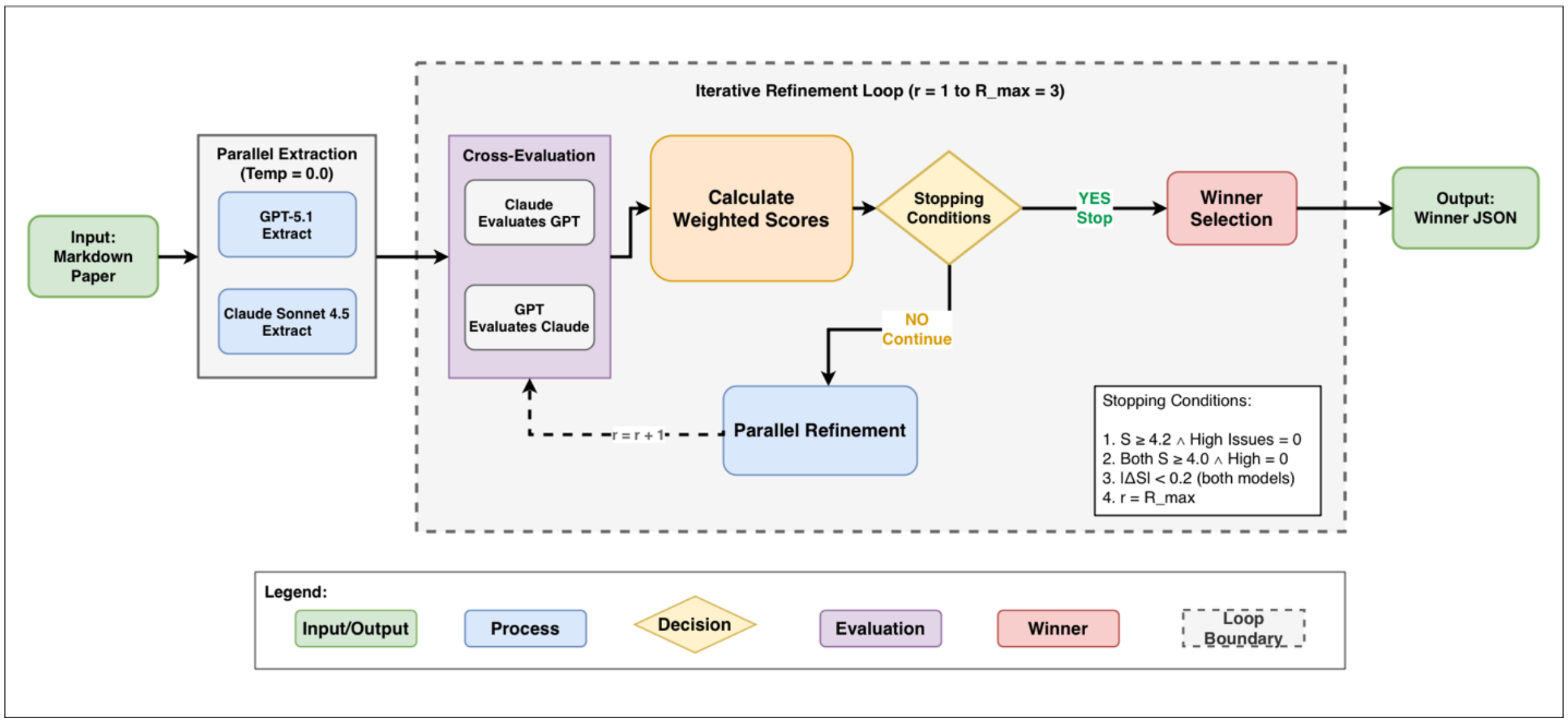

*Figure 2: Proposed Multi-Model Extraction Framework with Iterative Refinement*

Following the initial extraction, a comprehensive evaluation mechanism is implemented that extends the LLM-as-a-Judge (LaaJ) framework [34], which leverages the ability of LLMs to assess outputs generated by other models. Unlike conventional LaaJ implementations that typically rely on a single judge in a one-shot setting, an iterative cross-evaluation mechanism is introduced in which each model's output is assessed by the other model: Claude evaluates GPT's extraction, and GPT evaluates that of Claude. This symmetric cross-judging design provides two advantages, it mitigates systematic biases associated with self-evaluation or single-judge configurations [35], and it enables progressive refinement through feedback, with each model functioning as both extractor and evaluator across multiple rounds. The evaluating model receives only the JSON extraction output and paper identifier, without access to the original paper text, ensuring that evaluation focuses on internal consistency, schema completeness, and clinical plausibility [36,37].

The evaluation is organized across six core dimensions, each scored on a five-point scale. Table 3 summarizes the definition and focus of each evaluation dimension. Movement-related correctness is treated as a critical priority; any mistakes involving condition classification, contradictions in head movement fields, or implausible measurement values are designated as high-severity issues and result in dimension scores of 2 or lower. The evaluation output is reported as a structured record comprising individual dimension scores, an overall averaged score, a narrative justification, and an explicit count of high-severity issues.

*Table 3: Evaluation Dimensions and Scoring Criteria*

| Dimension | Definition |
|---|---|
| Completeness | Coverage of all patient groups with appropriate granularity |
| Quantitative Accuracy | Correctness of numerical values and accurate linkage to the corresponding patient groups |
| Symptom Extraction | Thoroughness and proper allocation of clinical symptoms |
| Head Movement Classification | Appropriate use of controlled vocabulary for describing movement characteristics |
| Schema Compliance | Adherence to structural and formatting requirements |
| Edge Case Handling | Conservative use of "NR" markers for unreported information |

An iterative refinement loop is introduced to improve extraction quality using targeted corrections derived from cross-evaluation feedback. In each round, structured feedback is provided to the corresponding model together with its previous extraction output. Models are instructed to trust quoted corrections, apply only justified changes, and preserve correct information. Refinement for both models is executed in parallel with temperature set to 0.0, followed by a new cross-evaluation cycle using the updated outputs. The process runs for a maximum of three rounds with early stopping when predefined criteria are met: one model achieves a weighted score of at least 4.2 with no high-severity issues, both models reach scores of at least 4.0 with no high-severity issues, or improvement between rounds falls below 0.2 points. This bounded iteration ensures computational efficiency while allowing meaningful improvements in extraction quality.

After refinement convergence or completion of the maximum number of rounds, a winner selection is applied to determine which model's extraction demonstrates superior quality for each paper. Rather than relying on unweighted averages, the selection process employs a weighted scoring scheme aligned with domain-specific priorities: head movement classification (25%), completeness (25%), quantitative accuracy (20%), symptom extraction (20%), schema compliance (5%), and edge case handling (5%). The model with the higher weighted score is designated as the winner, with a tie margin of 0.1 points used to resolve near-equivalent performance. The complete procedural logic is defined in Algorithm[1], which formalizes the dual-model extraction and refinement workflow, including scoring, stopping criteria, and final output selection.

**ALGORITHM 1: MULTI-MODEL EXTRACTION WITH CROSS-EVALUATION AND ITERATIVE REFINEMENT**

```
Input: Markdown paper text M; paper identifier P; maximum rounds R_max = 3
r ← 1
J_GPT ← ExtractWithGPT(M, P)                          //(initial GPT extraction)
J_Claude ← ExtractWithClaude(M, P)                    //(initial Claude extraction)
for rounds r = 1, 2, ..., R_max do
    E_GPT ← ClaudeEvaluates(J_GPT, P)                 //(cross-evaluation)
    E_Claude ← GPTEvaluates(J_Claude, P)
    S_GPT ← WeightedScore(E_GPT)
    S_Claude ← WeightedScore(E_Claude)
    if (S_GPT ⩾ 4.2 ∧ H(E_GPT)=0) or (S_Claude ⩾ 4.2 ∧ H(E_Claude)=0) then
        break (excellent extraction achieved)
    if (S_GPT ⩾ 4.0 ∧ S_Claude ⩾ 4.0) and H(E_GPT)=0 and H(E_Claude)=0 then
        break (both extractions acceptable)
    if minimal score improvement then
        break (convergence reached)
    J_GPT ← RefineWithGPT(J_GPT, E_GPT)
    J_Claude ← RefineWithClaude(J_Claude, E_Claude)
end
W_GPT ← FinalWeightedScore(E_GPT)
W_Claude ← FinalWeightedScore(E_Claude)
if |W_GPT − W_Claude| < 0.1 then
    winner ← argmax(OverallScore)                     //(tie-breaking)
else
    winner ← argmax(W_GPT, W_Claude)
J_winner ← extraction of winner model
Output: J_winner with confidence metadata
```

The dataset is organized into three main folders, as illustrated in Figure 3. The Abnormal Head Movements folder contains patient-group-level records describing AHMs, the Kinematics Quantitative folder contains quantitative measurements reported across the source publications, and the Severity Scales folder contains clinical severity scores extracted from heterogeneous rating instruments. Each folder is further subdivided at the disease and disorder levels, reflecting the classification scheme applied during the extraction pipeline.

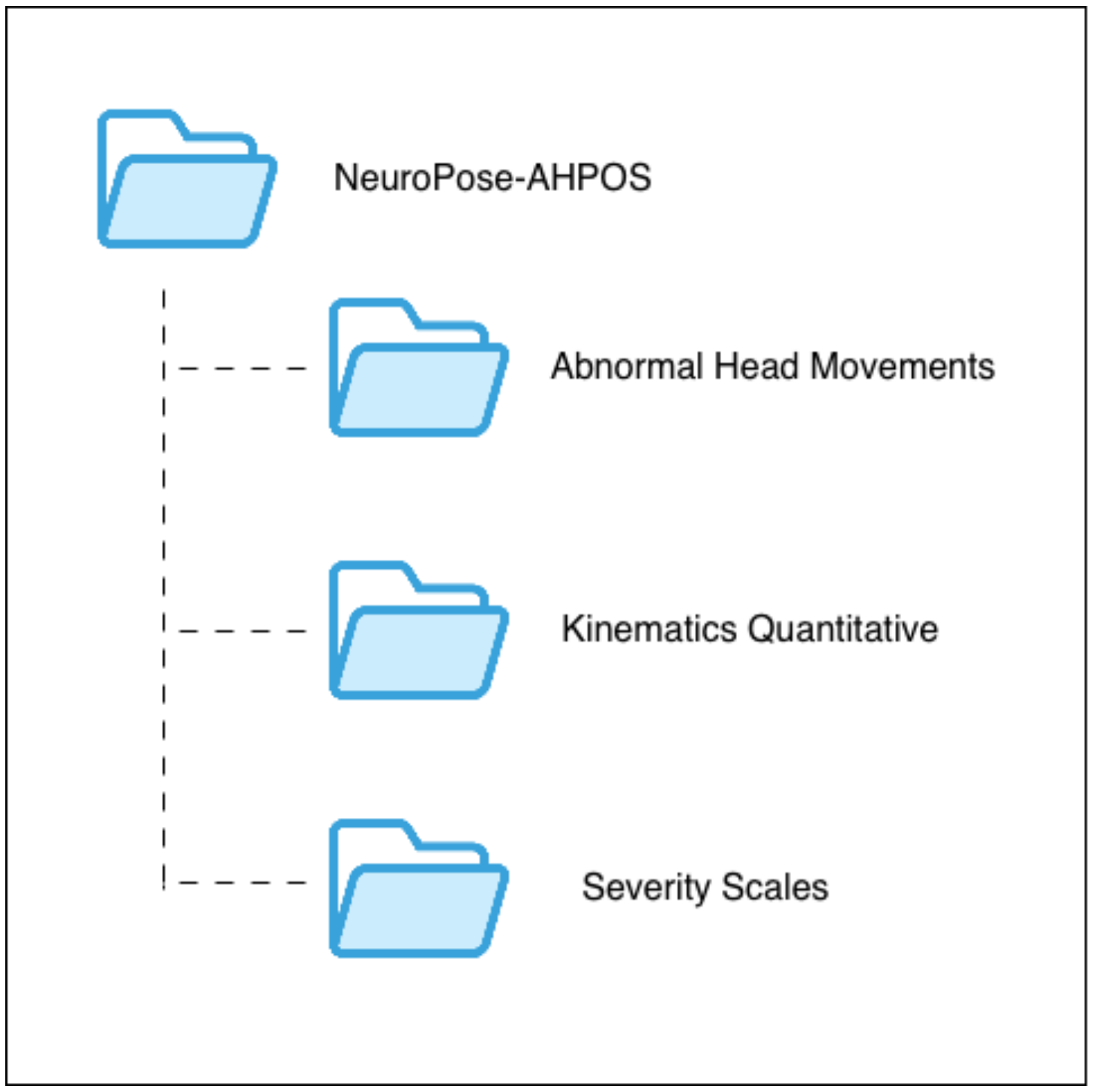

*Figure 3: NeuroPose-AHPOS Dataset Structure*

### *3.1.6 Dataset Quality Validation*

An inter-LLM agreement analysis is conducted to assess the reliability of AI-generated extractions for systematic review and structured data curation, providing empirical validation of the capacity of LLMs to extract clinical information from unstructured medical texts within the movement disorders domain. Moreover, a key consideration in this analysis is the recognition that some clinical terminology exhibits lexical variation when referring to the same underlying concept. For example, terms such as *anterocollis*, *forward head flexion*, and *head drop* may describe an identical movement pattern despite textual differences. This analysis incorporates semantic similarity measures rather than relying solely on exact categorical agreement to enable the differentiation between systematic bias and random variation while maintaining rigorous standards for evaluating extraction reliability and data quality. A set of statistical measures is employed to assess inter-LLM reliability, with the choice of metric determined by the data type. This analysis provides comprehensive coverage across categorical, continuous, set-based, and semantically variable fields. The analysis is conducted across all 1200 confirmed studies in the corpus, with each field evaluated over all papers for which that field is applicable, ensuring that agreement estimates reflect the full extraction pipeline.

### *A. Categorical Variables*

Cohen's kappa coefficient (κ) is utilized [38] to assess agreement for categorical fields including study type, condition classification, and head movement characteristics. It is selected because it corrects for chance agreement, providing a more conservative estimate of inter-LLM concordance than simple percent agreement. The kappa coefficient is defined as:

$$\kappa = \frac{(p_o - p_e)}{(1 - p_e)} \tag{1}$$

where $p_o$ is the observed proportion of agreement between the two models, and $p_e$ is the expected proportion of agreement by chance. The expected agreement is calculated as:

$$p_e = \sum_{i=1} p_{1i} \times p_{2i} \tag{2}$$

where $k$ is the number of categories, and $p_{1i}$ and $p_{2i}$ are the proportions of items assigned to category $i$ by model 1 (GPT 5.1) and model 2 (Claude Sonnet 4.5), respectively.

### *B. Continuous Variables*

For continuous data fields such as total sample size, group sizes, and quantitative head movement measurements (frequency, velocity, and amplitude), the Intraclass Correlation Coefficient (ICC) is calculated [39]. The ICC(2,1) model is used, representing a two-way random-effects model for absolute agreement. This approach treats both the models and the targets as random effects, allowing generalization beyond the sample. The ICC for absolute agreement is calculated as:

$$ICC(2,1) = \frac{(MS_R - MS_E)}{(MS_R + (k-1)MS_E + \frac{k}{n}(MS_C - MS_E)} \quad (3)$$

where $MS_R$ is the mean square for rows (subjects), $MS_E$is the mean square error (residual variance), $MS_C$ is the mean square for columns (models), k is the number of models (k=2), and n is the number of subjects.

*C. Set-based Variables*

The Jaccard similarity coefficient [40] is applied to set-based variables where multiple values could be extracted per field, specifically clinical symptoms (lists of symptoms per patient group) and clinical scale types (as multiple assessment scales may be reported for a single patient group). For two sets A and B, the Jaccard coefficient is defined as:

$$J(A,B) = \frac{|A \cap B|}{|A \cup B|} \quad (4)$$

where $|A \cap B|$ represents the number of elements common to both sets, and $|A \cup B|$ represents the total number of unique elements across both sets. The coefficient yields values ranging from 0 (no overlap) to 1 (complete overlap).

*D. Clinical Semantic Similarity*

Lastly, this component addresses the limitations of conventional string-based agreement metrics by enabling recognition of semantically equivalent terminologies. Rather than treating terminological differences as disagreements, this approach acknowledges the nuanced variability of clinical language, particularly in the classification of AHM. To overcome the limitations of exact string matching in evaluating model agreement, a domain-specific semantic similarity matrix is constructed.

A domain-specific semantic similarity matrix was constructed to quantify the clinical relatedness between pairs of movement descriptors. Each entry in the matrix reflects expert-informed similarity scores based on shared pathophysiology, common diagnostic usage, or clinical co-occurrence patterns. For instance, *head drop* and *forward flexion* are assigned a high similarity score ($S = 0.8$) due to their near-equivalent depiction of anterior displacement. Similarly, *anterocollis* and *head drop* are rated as closely related ($S = 0.7$), reflecting overlapping pathophysiological features, while *cervical dystonia* and *torticollis* receive a score of $S = 0.6$ to reflect a taxonomic relationship, where the latter represents a specific manifestation of the broader condition.

Given two sets of extracted movement types, $M_{\mathrm{GPT}}$ and $M_{\mathrm{Claude}}$, the maximum semantic similarity between non-overlapping terms is computed as:

$$S_{\max} = \max_{\{m_i \in M_{GPT} - M_{Claude}, m_j \in M_{Claude} - M_{GPT}\}} S(m_i, m_j) \quad (5)$$

This measure identifies the highest degree of clinical correspondence between non-overlapping subsets of extracted terms which allows the similarity metric to account for medically relevant terminologies.

### *3.2 Cervical Dystonia Analysis: Movement Classification, Severity Indexing, and Clinical Validation*

This section focuses on the characterization and benchmarking of CD using structured clinical data extracted from NeuroPose-AHM dataset. The proposed methodology comprises four components, as illustrated in Figure 4. Task 1 addresses the classification of AHM types from quantitative kinematic measurements. Task 2 constructs a unified literature-derived severity index, the Head–Neck Severity Index (HNSI), by unifying heterogeneous clinical rating scales into a normalized metric. Task 3 validates the HNSI normalization against a real-world patient dataset to assess clinical alignment through statistical benchmarking. Task 4 performs a bridge analysis by correlating the movement-type probability outputs from Task 1 with the HNSI scores from Task 2 across a shared subset of studies, providing an assessment of the framework's internal consistency.

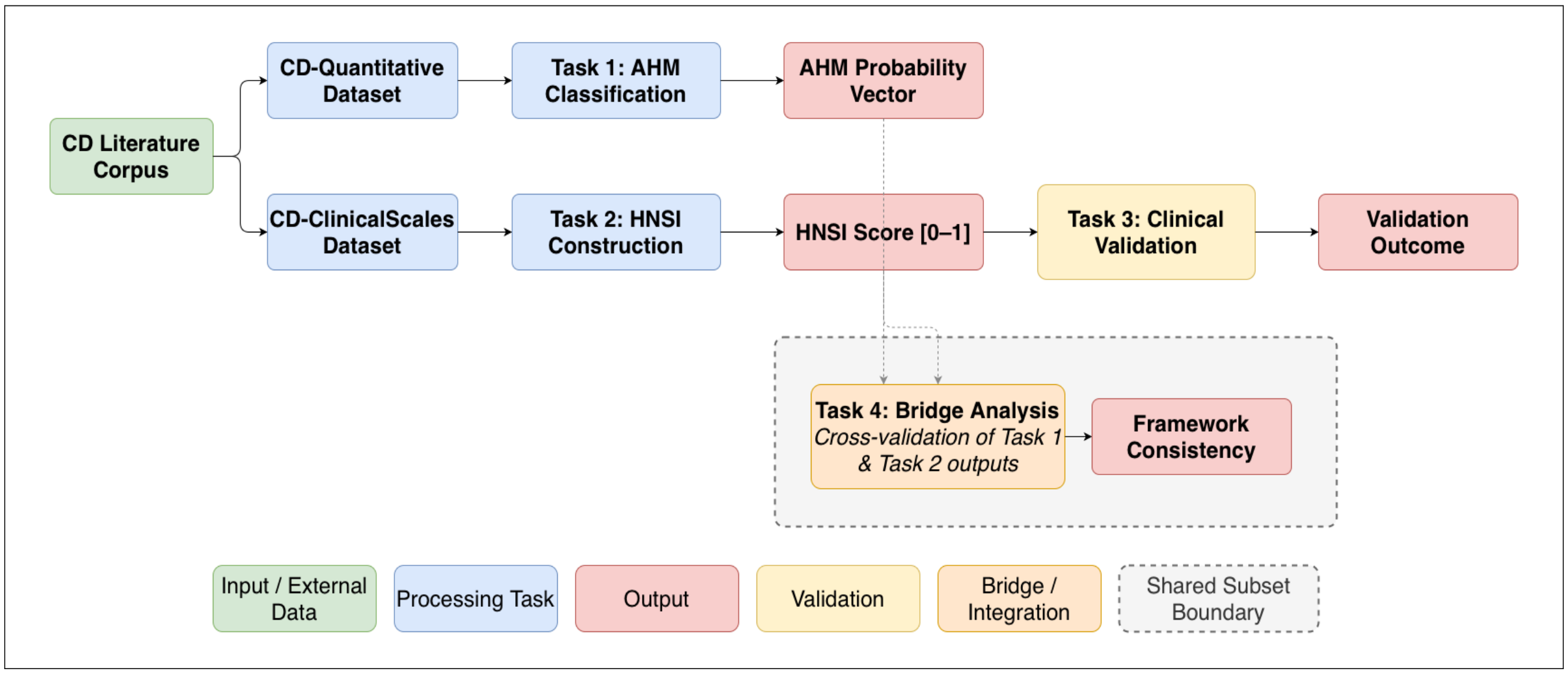

*Figure 4: Cervical Dystonia Analytical Framework*

### *3.2.1 Cervical Dystonia Subset Selection*

Following the construction of the NeuroPose-AHM dataset, the present analysis focuses specifically on records associated with Cervical Dystonia. Among 846 collected papers, the full dataset is filtered to retain only records for CD condition, comprising 202 papers. The resulting CD specific records are subsequently partitioned into two functionally distinct datasets according to the data type they contain. The first partition, designated the CD-Quantitative (CD-Q) dataset, aggregates records containing quantitative kinematic measurements, including frequency, velocity, and amplitude values, associated with their measurement systems. This dataset comprises 113 records derived from 45 papers and constitutes the primary input for the movement classification task. The second partition, designated the CD-Clinical Scales (CD-CS) dataset, including records containing clinical scale assessments, including scale names, subscale identifiers, score ranges, baseline values, post-treatment values, and associated statistical metrics. This dataset comprises 809 records across 11 columns derived from 137 papers and serves as the source from which the four target scales for HNSI construction are extracted. The dataset, while focused in scope, is grounded in expert-level clinical knowledge extracted from peer-reviewed neurological literature; therefore, the feature space reflects clinically validated assessment protocols rather than purely data-driven abstractions.

### *3.2.2 Task 1: Abnormal Head Movement Classification*

This task addresses the classification of AHM types in CD patients based on quantitative kinematic measurements including movement frequency, peak velocity, amplitude, and latency. CD presentations are characterized by the frequent co-occurrence of multiple movement types. Rather than manifesting a single isolated pattern, a substantial proportion of patients exhibit concurrent combinations, such as torticollis with laterocollis or anterocollis with head tremor. Accurate identification of the specific movement types present and their combinations, has direct diagnostic and therapeutic relevance as the pattern of muscular involvement influences both the clinical severity profile [41] and the selection of target muscles for botulinum toxin injection therapy [42,43]. Within the broader analytical framework of this research, movement analysis is formulated as a supervised learning problem, using quantitative kinematic measurements extracted from the published CD literature as predictive inputs. The resulting binary prediction vectors and class probability outputs further serve as kinematic inputs to the bridge analysis described in Task 4, where movement profiles are cross-validated against derived clinical severity estimates.

### *3.2.2.1 Data Preprocessing*

The CD-Q dataset is derived from heterogeneous clinical studies, each employing distinct measurement conventions and reporting standards. As a result, the raw extracted records exhibit variability in completeness, terminology, and unit representation. To mitigate these cross-study inconsistencies and prepare the data for supervised learning, a structured preprocessing framework is implemented, comprising data filtering, label standardization, and feature engineering. These steps transform the raw extracted records into a structured, model-ready representation suitable for supervised multi-label learning.

#### *A. Data Filtering*

A data quality filter is applied in which records with all four kinematic measurement fields (amplitude value, frequency value, latency value, and velocity value) are simultaneously absent are excluded. Such records represent papers that reported a movement type label without providing any accompanying quantitative kinematic measurement. Including these records would compromise the integrity of the training process, leading the model to infer movement classifications from statistical noise rather than authentic kinematic patterns. Following the application of this filter, the dataset is reduced from 153 rows across 58 papers to 113 rows across 45 papers, which constitutes the definitive input dataset for AHM classification.

### *B. Data Standardization*

Following data filtering, a standardization process is applied to the collected AHM labels, as different terminologies are used inconsistently across studies to describe clinically equivalent presentations. To ensure consistency, this step unifies alternative terms referring to the same movement type. Terminology is standardized for two AHM categories: “torticollis,” which may be described as “dystonic,” “dystonia,” “rotational CD,” or “spasmodic torticollis”; and “head tremor,” which may be referred to as “oscillation” or “jerky” movements. The resulting label distribution across the 113 records comprises five AHM type labels as presented in Figure 5. Torticollis emerges as the predominant AHM type, substantially exceeding the frequency of other movement categories. Head tremor and laterocollis follow at moderate prevalence, whereas anterocollis and retrocollis appear comparatively less frequently.

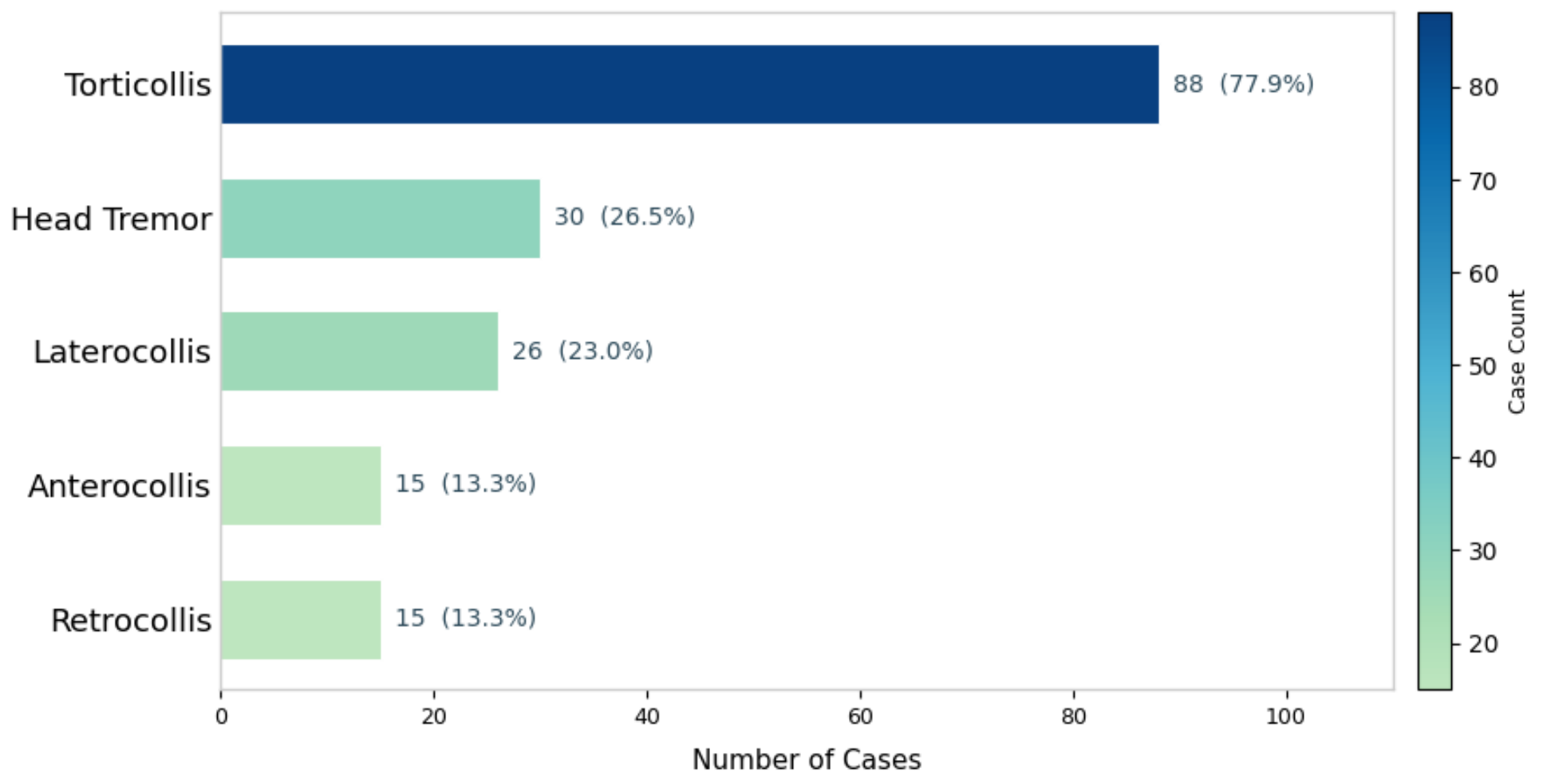

*Figure 5: AHM Labels Distribution*

To further examine label interactions, Figure 6 presents an AHM multi-label combinations summarizing all unique co-occurrence patterns. The largest intersection corresponds to torticollis as an isolated presentation, confirming its dominance even when label combinations are considered. Among multi-label patterns, torticollis most frequently co-occurs with laterocollis, followed by pairings with anterocollis and retrocollis. Higher-order combinations involving three or more movement types are present but relatively rare.

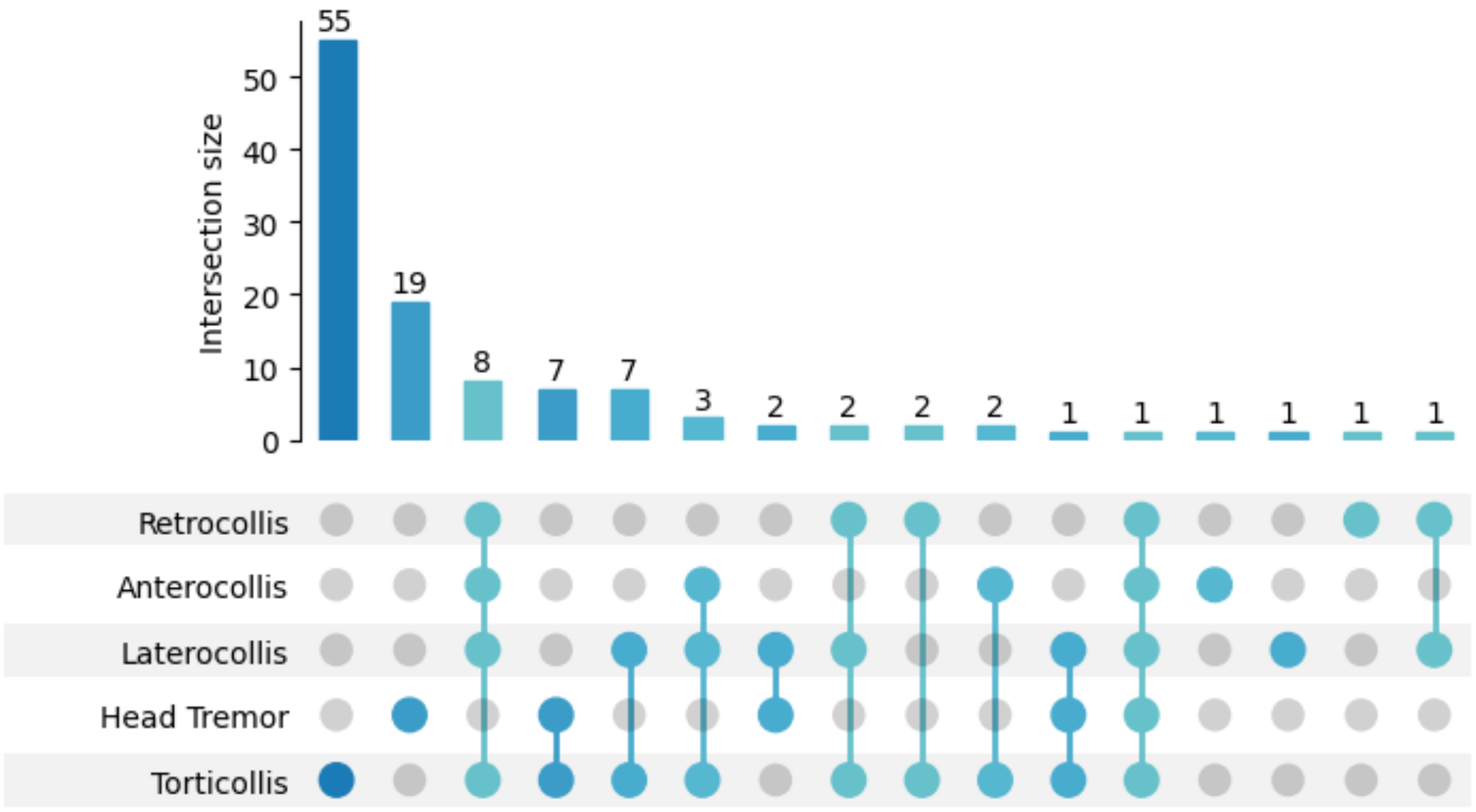

*Figure 6: AHM Multi-Label Combinations*

## *C. Features Engineering*

Five feature categories are constructed from the raw kinematic columns to produce a structured representation. Each category addresses a distinct challenge arising from the heterogeneity of measurement units, acquisition modalities, and physiological plausibility constraints encountered in the literature-derived data.

1. Amplitude normalization unifies heterogeneous reporting units (degrees, millimetres, centimetres, electromyography (EMG) percentage, dimensionless values) through unit-aware conversion. Centimetre values are converted to millimetres, while angular and normalized values are retained as reported. Missing values are imputed using within-class AHM-type medians to preserve intra-class kinematic structure.
2. Frequency cleaning applies physiologically grounded range constraints corresponding to resting tremor (3-6 Hz), action tremor (4-11 Hz), and dystonic contractions (1-6 Hz) [44,45]. Values outside these ranges are set to missing and subsequently imputed using per-movement-type medians.

3. Latency cleaning retains only millisecond-unit records, converting second-unit values through multiplication by 1000, followed by per-movement-type median imputation of residual missing values.
4. Measurement system encoding converts acquisition modalities including EMG, inertial measurement units (IMU), optical motion capture systems, magnetic resonance imaging (MRI), and related measurement systems into ordinal categories. Modality is retained as a covariate due to its influence on noise characteristics and temporal and spatial resolution across studies.
5. Amplitude unit type encoding applies one-hot encoding to amplitude unit categories (degrees, millimetres, centimetres), generating indicator columns that allow the classifier to account for the measurement unit when learning amplitude-class relationships.

#### *3.2.2.2 Multi-Label Classification Framework*

The AHM prediction task is implemented as a supervised multi-label classification problem. The aim is to learn a mapping between structured kinematic features and the concurrent presence of one or more AHM types. The feature space consists of clinically tabular variables, a domain in which traditional ML classifiers have been shown to match or outperform DL models on small-to-medium scale datasets [46,47]. Therefore, five ML baseline classifiers are evaluated: Logistic Regression (LR) [48], Random Forest (RF) [49], Extreme Gradient Boosting (XGBoost) [50], Light Gradient Boosting Machine (LightGBM) [51], and a shallow Multilayer Perceptron (MLP) [52]. All classifiers are trained and evaluated using stratified five-fold cross-validation, implemented at the label-set level to preserve the distribution of multi-label combinations across folds.

### *3.2.3 Task 2: Head-Neck Severity Index Construction*

The clinical assessment of CD relies on structured severity assessment to guide diagnosis, treatment planning, and longitudinal monitoring. In both research and clinical practice, severity is measured using several validated rating instruments including the TWSTRS, Tsui Scale, TRS, and GDRS. Each one of these scales is defined by distinct scoring ranges, subscale structures, and clinical alignment frameworks. As a consequence of this heterogeneity, raw scale scores cannot be directly compared or aggregated across studies. Moreover, no individual instrument offers comprehensive representation across all 137 papers comprising the extracted CD-CS dataset. Therefore, this task addresses the construction of a unified severity representation through the unifies of these heterogeneous scales. The resulting HNSI yields a single normalized score within the interval [0, 1] for each paper, where 0 denotes the absence of clinically significant impairment and 1 corresponds to the maximum recorded severity under the most restrictive scale. Within the overall framework, the index serves both as the quantitative basis for the clinical validation in task 3 and as one of the key outputs examined in the bridge analysis in Task 4.

### *3.2.3.1 Clinical Context and Scale Selection*

Clinical severity in CD is assessed using validated rating scales administered by clinicians. Given the structural differences among these instruments, direct comparison of raw scores across scales is inappropriate. For instance, a TWSTRS severity score of 20 out of 35 and a Tsui score of 10 out of 20 both represent approximately 57% of maximum severity, but this equivalence is only interpretable after scale-aware normalisation. Four scales represented in the CD-CS dataset are selected for HNSI construction, based on their coverage of the head and neck region. Table 4 summarizes each scale, its head/neck subscale scope, and the maximum score used as the normalisation denominator.

*Table 4: Clinical Scales Selected for HNSI Construction*

| Scale | Head/Neck Subscale | Max Score | Scope |
|---|---|---|---|
| TWSTRS | Severity Scale (TSS) | 35 | CD specific |
| Tsui | Total score | 20 | |
| TRS | Head/Face subscale | 8 | |
| GDRS | Neck subscale | 10 | Generalized dystonia |

### *3.2.3.2 Scale Extraction and Normalization*

For each selected instrument, a subscale filtering procedure is applied to the CD-CS dataset to identify records eligible for HNSI computation. TWSTRS records are retained when the subscale field contains any of the following keywords: severity scale, TSS, severity of torticollis, overall severity, torticollis score, laterocollis score, retrocollis score, or shoulder displacement.

In contrast, Tsui scale records are included without subscale restriction, as the entire instrument is specific to head and neck involvement. TRS and GDRS records are retained when the subscale field contains the keywords head, face, neck, or cranial. For the score selection, baseline scores are used as the primary basis for normalization. Records with missing baseline values are excluded from analysis. Following application of these criteria, the resulting dataset comprises 278 rows derived from 66 papers.

The HNSI is computed through a two-stage aggregation procedure. In the first stage, each retained clinical score is normalized to the [0, 1] interval using the scale-specific head-neck maximum ($hn_max_k$), where $hn_max_k$ denotes the maximum possible score for the head and neck subscale of scale k (35 for TWSTRS, 20 for Tsui, 8 for TRS, and 10 for GDRS). A clipping operation is first applied to constrain any out-of-range values to the valid scoring interval, after which the score is divided by $hn_max_k$ to produce a normalized value. For cohort observation i under scale k, the normalized score is computed as:

$$\widetilde{s_{i,k}} = \frac{min\left(max\left(s_{i,k}|0\right)|hn_max_k\right)}{hn_max_k}$$

In the second stage, where paper $p$ contributes $n_{p,k}$ observations from scale $k$, a scale-level score is computed as the arithmetic mean of all normalized observations for that paper under that scale:

$$\overline{s_{p,k}} = \frac{1}{n_{p,k}}\sum_{i=1}^{n_{p,k}}\widetilde{s_{i,k}}$$

The final HNSI for paper $p$ is then the arithmetic mean across all $K_p$ scales represented in that paper, where $K_p \in \{1, 2, 3, 4\}$ denotes the number of distinct scales contributing observations for paper p:

$$HNSI_p = \frac{1}{K_p}\sum_{k=1}^{K_p}\overline{s_{p,k}}$$

This produces a single HNSI value per paper on the [0, 1] interval, where 0 represents the absence of clinically significant head-neck impairment and 1 represents the maximum severity recorded on the most restrictive scale.

### *3.2.4 Task 3: Clinical Validation Against Real-World Patient Data*

The earlier components of this framework— the AHM multi-label classifier (Task 1) and the Head-Neck Severity Index (Task 2) — are derived from group-level statistics aggregated from literature papers in the NeuroPose-AHM dataset. While this design establishes methodological coherence within the framework, it needs more validation on observable clinical patterns in individual patients. To evaluate the clinical grounding of the proposed framework, Task 3 performs an external proof-of-concept clinical validation of both the AHM multi-label classifier and the HNSI against an independent real-world patient cohort.

The validation dataset is derived from the prospective cohort reported by Nakamura et al. (2019), conducted at Juntendo University Hospital, Tokyo [23]. The study enrolled 30 consecutive patients diagnosed with CD who were undergoing botulinum toxin therapy. Severity assessment was performed using a Kinect v2–based semi-automated motion capture system that quantified head posture across three axes and integrated these measurements into the computation of the TWSTRS severity subscale score. In parallel, a movement disorder–trained neurologist independently rated the same patients based on video recordings. Table 5 presents the characteristics of the validation dataset.

*Table 5: Characteristics of Validation Dataset*

| Characteristic | Value |
|---|---|
| Age (years), mean ± SD | 52.3 ± 16.0 |
| Sex | 18 (60%) male<br>12 (40%) female |
| Disease duration (years), mean ± SD | 11.3 ± 12.7 |
| AHM | Torticollis (n=26)<br>Laterocollis (n=26)<br>Anterocollis (n=5)<br>Retrocollis (n=12) |
| TWSTRS severity score, mean ± SD | 18.8 ± 3.9 |
| Normalized severity range | 0.229–0.800 |
| Assessment modality | Kinect v2 semi-automated motion capture |

#### *3.2.5 Task 4: Bridge Analysis and Cross-Component Validation*

The final task in this framework performs an internal cross-task validation by examining the coherence between the primary outputs of Task 1 (movement type classification) and Task 2 (Head-Neck Severity Index construction). The conceptual premise underlying this analysis is that, if Task 1 successfully captures the kinematic movement profile of CD and Task 2 accurately quantifies clinical severity, then papers characterized by more complex movement patterns should also exhibit higher severity scores. This relationship constitutes an empirical hypothesis that can be evaluated only through joint analysis of the two derived measures.

This cross-task evaluation is enabled by the subset of 24 papers that contribute records to both the CD-Q and CD-CS datasets, yielding a bridge dataset of 68 linked observations for which both movement probabilities and HNSI scores are available. For each overlapping paper, the classifier trained in Task 1 generates a probability vector representing the predicted likelihood of each AHM type. These probabilities are treated as continuous indicators of movement pattern severity. The corresponding HNSI scores derived in Task 2 provide a normalized estimate of clinical severity for the same analytical units. Pairwise Pearson correlation coefficients are computed between HNSI scores and the predicted probability of each individual movement type, as well as a composite mean probability score defined as the mean of predicted probabilities across all movement types, serving as an aggregate indicator of the overall kinematic pattern expression for each paper. This composite metric serves as an aggregate representation of kinematic complexity and is hypothesized to exhibit a positive association with clinical severity.

The bridge analysis is used as an internal consistency assessment rather than a causal inference procedure, given the observational structure of the data and the modest size of the overlapping subset. Evidence of a coherent positive association between movement complexity and HNSI would support the construct validity of the framework, indicating that the two analytically distinct tasks capture complementary dimensions of the same underlying clinical phenomenon.

## 4. Results and Discussion

### *4.1 Dataset Extraction Quality*

The results are presented in two parts: first, characterization of the NeuroPose-AHM dataset across corpus, movement, and clinical scale dimensions; second, analytical benchmarking results for cervical dystonia across the four proposed tasks.

#### *4.1.1 Analysis Results*

As shown in Table 6, the inter-LLM agreement analysis demonstrates high overall reliability across study-level, patient group–level, movement-related, and quantitative fields. Per-field agreement values are reported for all schema elements across the full extraction corpus. At the study level, inter-LLM agreement was particularly strong, with study type classification achieving high agreement (κ = 0.822; 85.1% exact matches). Total sample size extraction exhibited extremely high reliability (ICC = 1.000), with exact agreement in 94.5% of papers and agreement within ±1 patient in 94.8% of cases.

At the patient group level, agreement varied according to extraction complexity and linguistic ambiguity. Condition name extraction demonstrated reliable agreement (κ = 0.709), indicating consistent identification of clinical conditions despite lexical variability in medical terminology. Condition category and causes of AHM achieved substantial agreement (κ = 0.663 and κ = 0.650, respectively), while patient count extraction showed strong reliability (ICC = 0.984). In contrast, symptom list extraction exhibited lower agreement, reflecting differences in granularity inherent to free-text symptom descriptions rather than substantive disagreement.

Lastly, movement type classification showed notable improvement with semantic similarity adjustment. Exact matching yielded moderate agreement (κ = 0.506), whereas similarity-adjusted analysis increased agreement to 78.4% (κ = 0.567), with a mean similarity score of 0.723. Other movement characteristics, including direction, laterality, consistency, and pattern, demonstrated substantial agreement without semantic adjustment. For studies reporting quantitative measurements, numerical values exhibited very high reliability for frequency (ICC

= 1.000), velocity (ICC = 0.951), and amplitude (ICC = 0.939), with minimal systematic bias, while identification of clinical assessment scales achieved good agreement (mean Jaccard = 0.605). Overall, across the analyzed fields, all achieved at least moderate reliability, and 79% demonstrated good to very high agreement, supporting the robustness of LLM data extraction in movement disorder systematic reviews.

*Table 6: Inter-LLM agreement metrics across fields*

| Category | Field | Measure | Value |
|---|---|---|---|
| Study Level | Study type | Cohen's κ | 0.822 |
| | Total sample size | ICC | 1.000 |
| Patient Groups | Condition name | Cohen's κ | 0.709 |
| | Condition category | | 0.663 |
| | Number of patients | ICC | 0.984 |
| | Causes AHM | Cohen's κ | 0.650 |
| | Head symptoms | Jaccard | 0.292 |
| | General symptoms | | 0.266 |
| Head Movements | Movement type | Cohen's κ | 0.567 |
| | Direction | | 0.675 |
| | Laterality | | 0.635 |
| | Consistency | | 0.761 |
| | Pattern | | 0.658 |
| Quantitative | Measurement performed | Cohen's κ | 0.558 |
| | Frequency | ICC | 1.000 |
| | Velocity | | 0.951 |
| | Amplitude | | 0.939 |
| Clinical Scales | Scale types | Jaccard | 0.605 |

### 4.1.2 *Winner Models Analysis*

A comparative analysis of average evaluation scores for GPT-5.1 and Claude Sonnet-4.5 across multiple diagnostic dimensions is presented in Figure 7. GPT achieves slightly higher average scores across all categories. The largest differences appear in completeness (4.76 vs. 4.13) and symptom extraction (4.64 vs. 4.01), indicating stronger coverage and clinical detail in GPT-generated outputs. Smaller differences are observed in head movement classification (4.55 vs. 4.45) and edge case handling (4.91 vs. 4.43), suggesting modest improvements in handling complex scenarios. Both models achieve identical performance in schema compliance, each reaching the maximum score of 5.00. These results indicate broadly similar performance across the evaluated diagnostic dimensions, with GPT showing modest advantages in several categories rather than consistent superiority across all evaluation criteria.

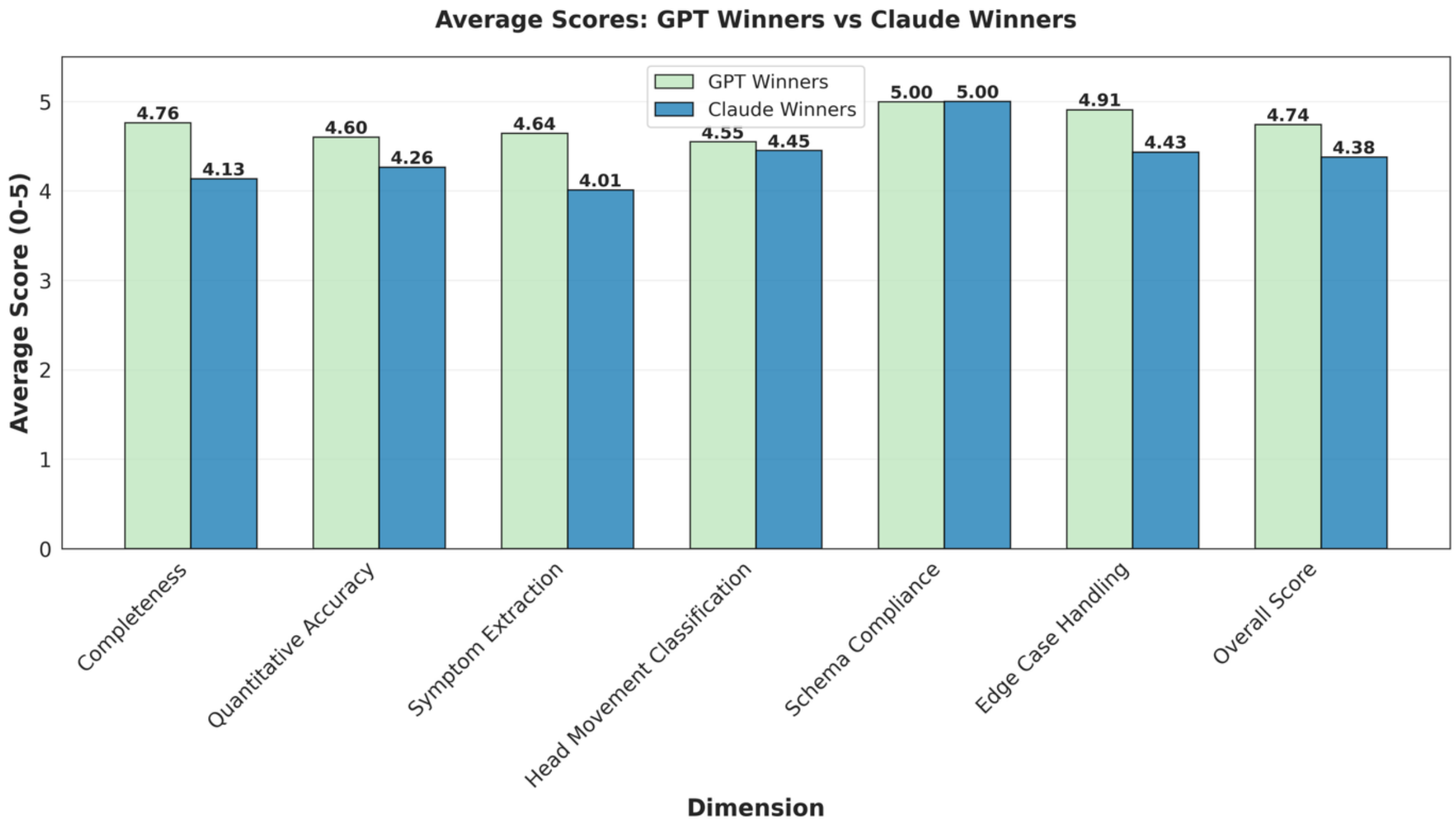

Figure 7: Average Score by Dimension

Figure 8 summarizes the overall distribution of winning outputs by model. The pie chart shows that GPT-5.1 accounts for 64.0% of winning generations, while Claude Sonnet-4.5 accounts for 36.0%. These results indicate that GPT-generated outputs are preferred more frequently across the evaluated criteria. The winning outputs from this evaluation constitute the final extracted records in NeuroPose-AHM. Following post-extraction filtering to retain only papers reporting AHM-related neurological disorders and diseases, 846 papers were included in the final dataset, yielding 2756 patient-group-level records.

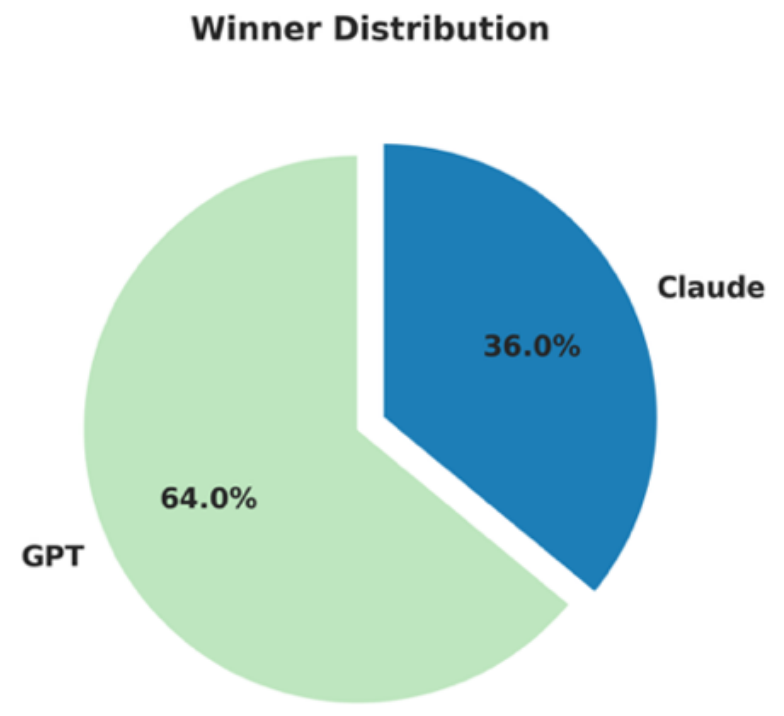

*Figure 8: Winners Distribution*

### *4.2 Literature Corpus Characteristics*

The collected papers provide insights into how various AHMs are characterized and analyzed in relation to neurological conditions and movement patterns. Figure 9 presents a co-occurrence network of the reviewed literature, illustrating how AHM-related terms such as torticollis, anterocollis, and retrocollis cluster at the intersection of neurological condition characterization.

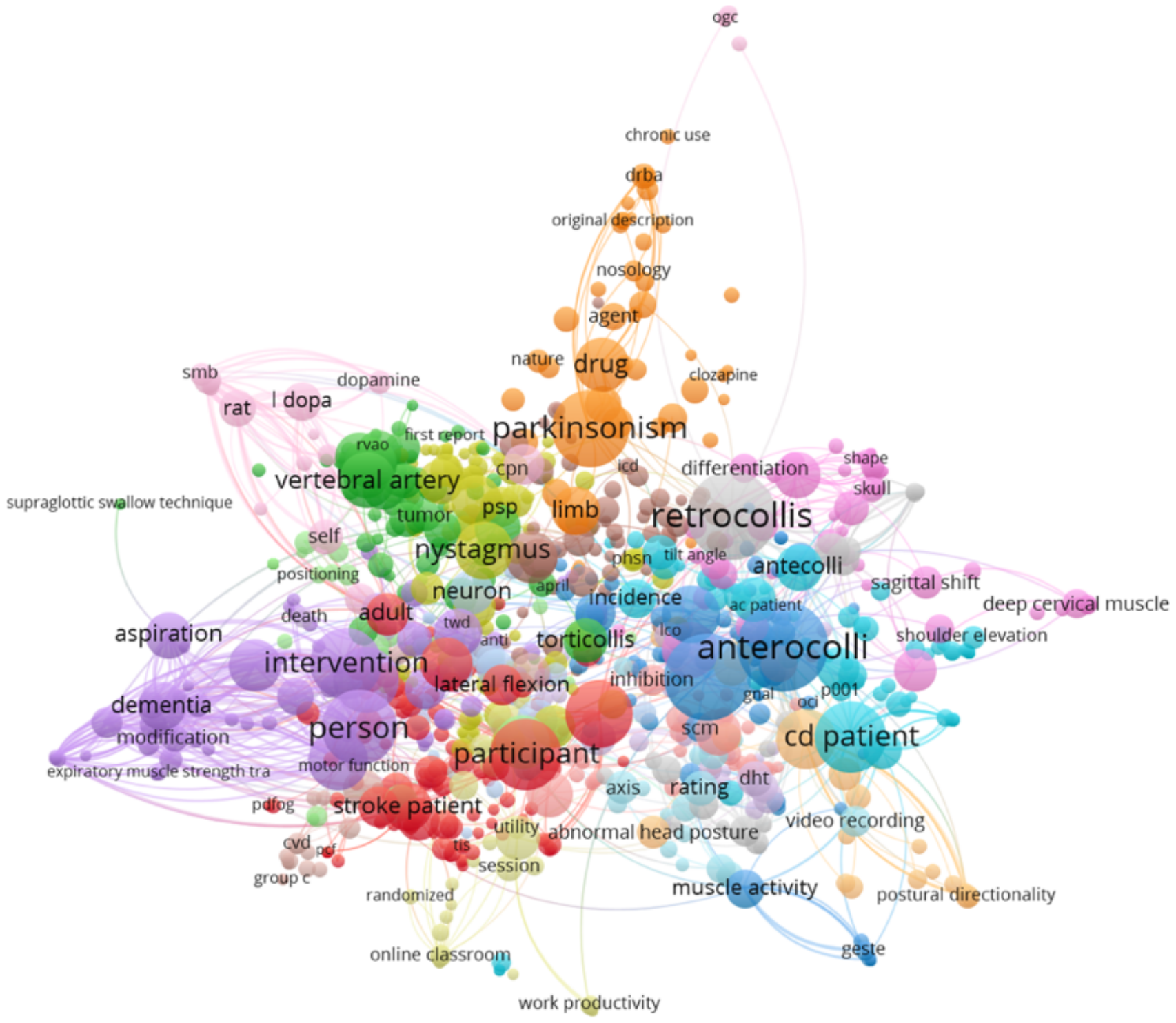

*Figure 9: Keyword co-occurrence network of AHM-related literature, generated using VOSviewer.*

The selected studies provide detailed insights into how AHM are clinically expressed across a range of neurological conditions. Figure 10 maps neurological conditions to specific AHM subtypes, with torticollis, laterocollis, anterocollis, and retrocollis emerging as the most frequently documented

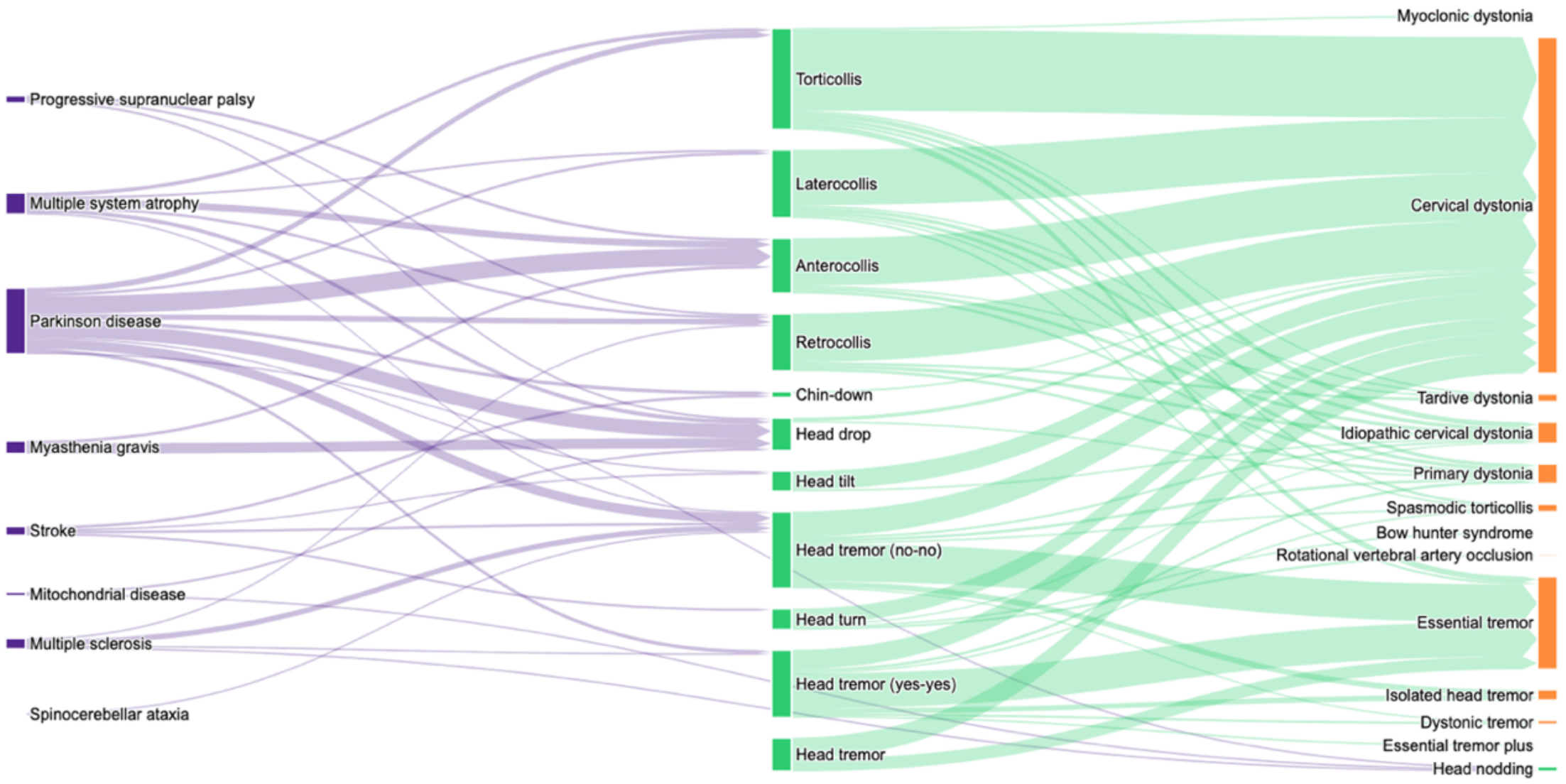

*Figure 10: Relationships between neurological conditions (left), abnormal head movements (center), and specific clinical manifestations (right).*

The distribution of study designs across the included literature is presented in Figure 11. Cross-sectional studies are the most prevalent, followed by case reports, case series, and prospective studies. Reviews and retrospective studies are also well represented, whereas cohort studies and randomized controlled trials (RCTs) are less frequent.

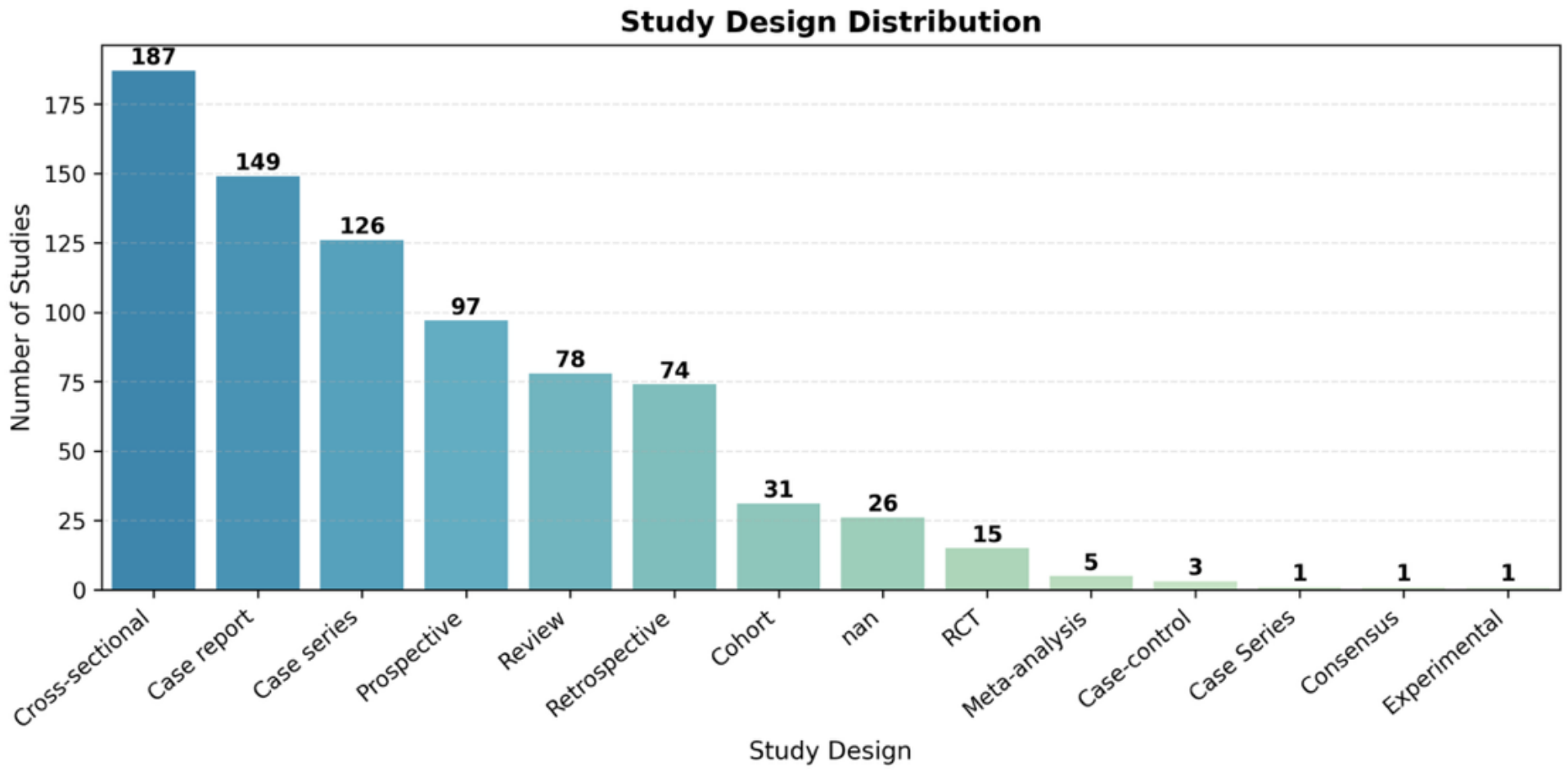

*Figure 11: Distribution of study designs among included AHM-related publications*

## *4.3 Neurological Conditions and Associated Movement Profiles*

To better understand the prevalence of AHM types in the literature , Figure 12 summarizes the distribution of the most frequently reported AHM subtypes, with head tremor emerging as the most commonly documented movement, followed by combined movement patterns, torticollis, anterocollis, and retrocollis.

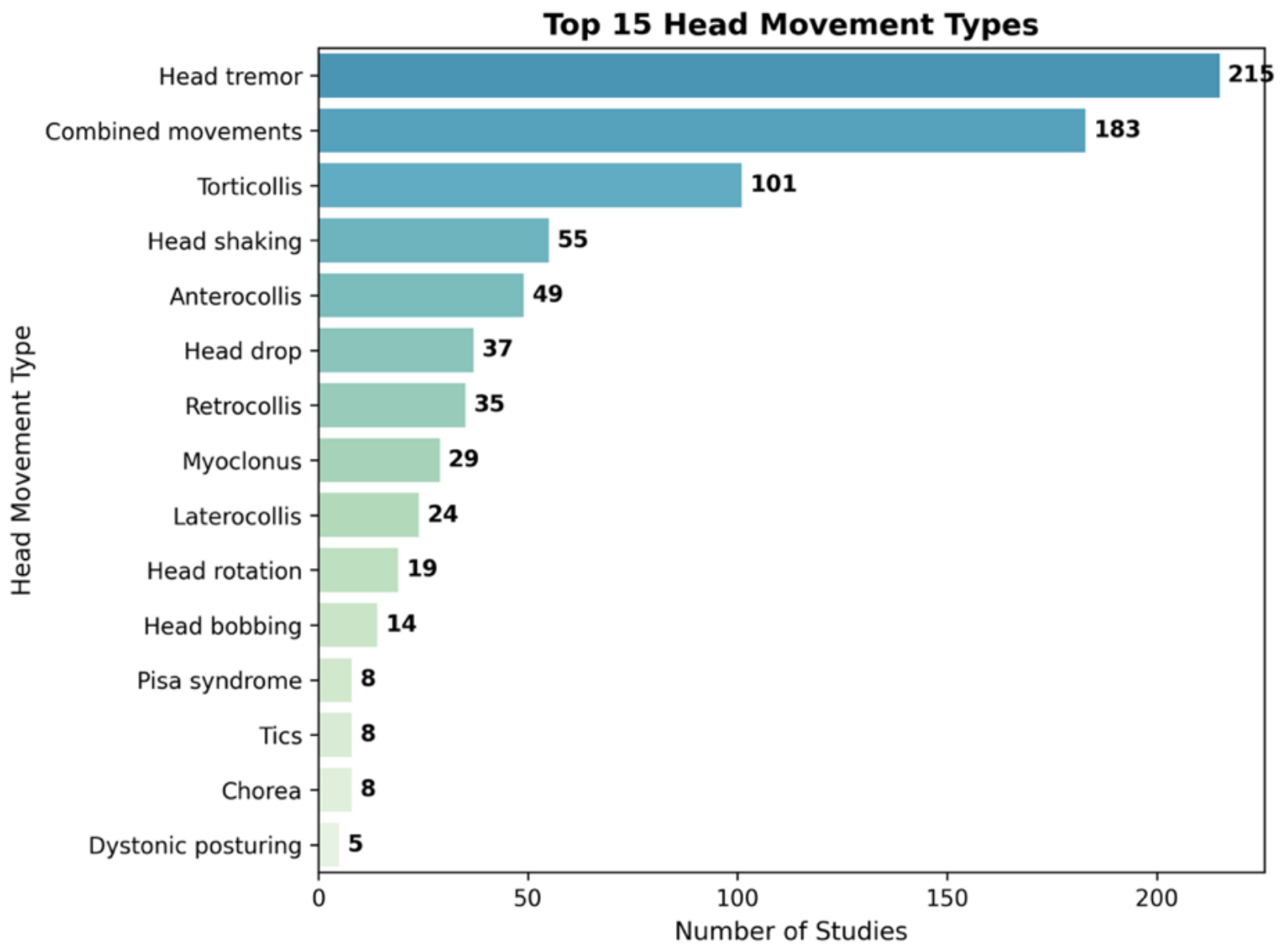

*Figure 12: Frequency distribution of the top AHM types.*

Figure 13 presents the most frequently reported disorders and diseases associated with AHM in the literature. Cervical dystonia is the predominant disorder, followed by essential tremor, underscoring the central role of movement-related conditions. At the disease level, Parkinson’s disease is most commonly represented, with stroke and multiple sclerosis also reported with notable frequency.

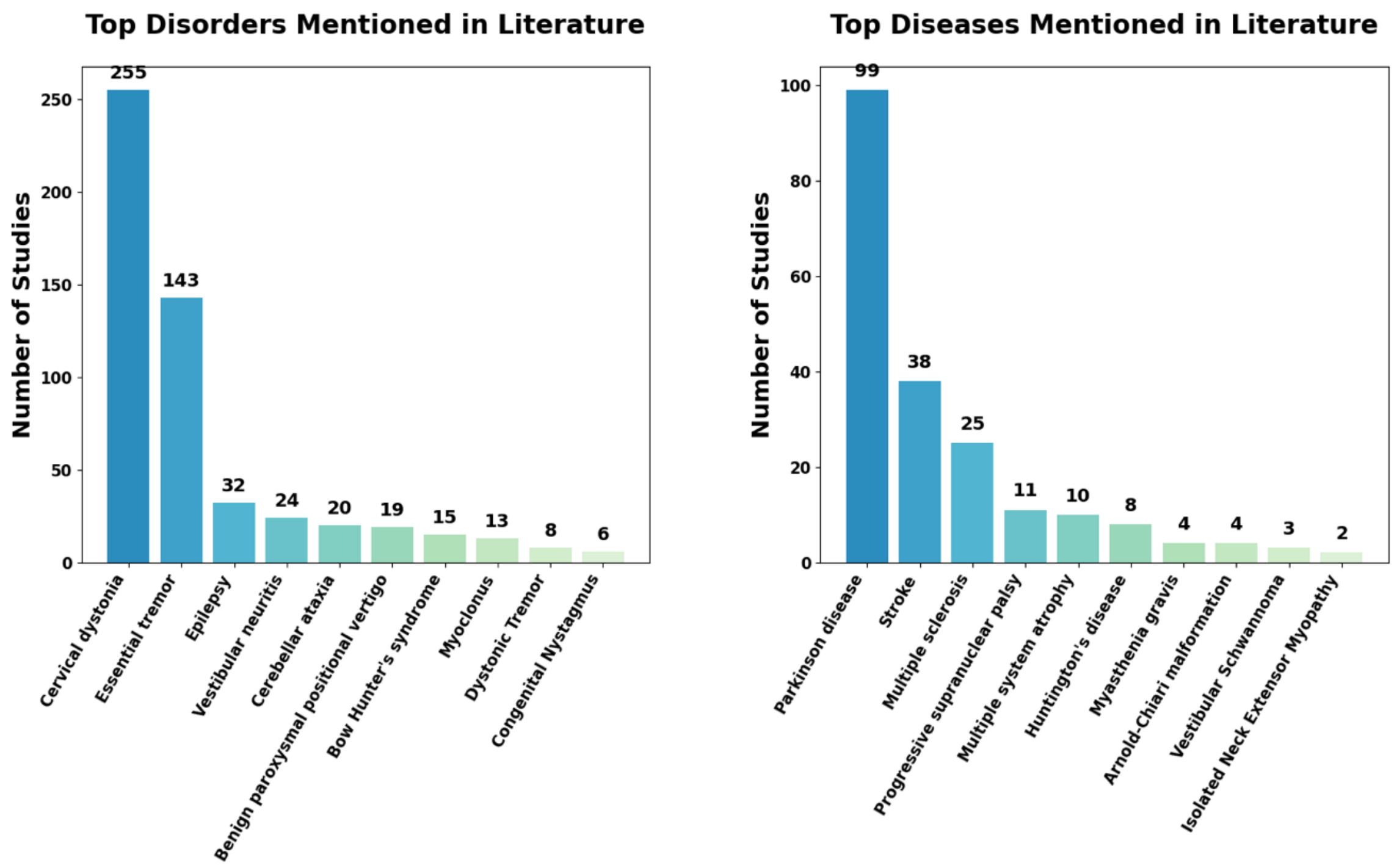

*Figure 13: Top Conditions Mentioned in Literature (a) Disorders (b) Diseases.*

Figure 14 and Figure 15 presents the characteristic symptom profiles for a subset of AHM-related disorders and diseases. Distinct differences in dominant head movement patterns are observed across conditions. Cervical dystonia is primarily characterized by combined movement patterns and torticollis, while cerebellar ataxia and epilepsy are more frequently associated with tremor- and myoclonus-related manifestations. At the disease level, Parkinson's disease and multiple sclerosis exhibit a broader distribution of movement types, whereas Huntington's disease and stroke are dominated by more specific movement patterns. Overall, these profiles illustrate condition-specific associations and AHM subtypes.

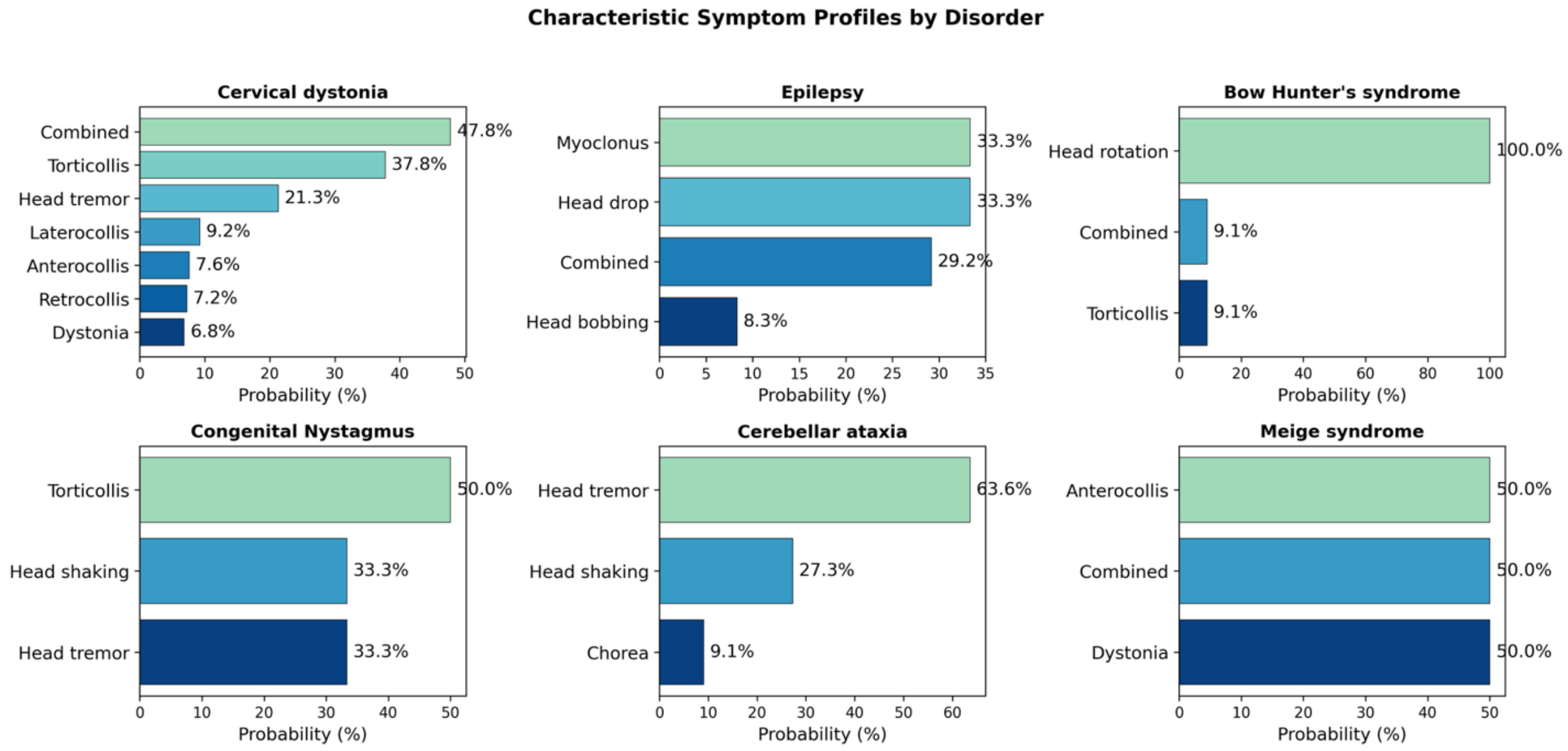

*Figure 14: Characteristic Symptom Profiles by Disorder*

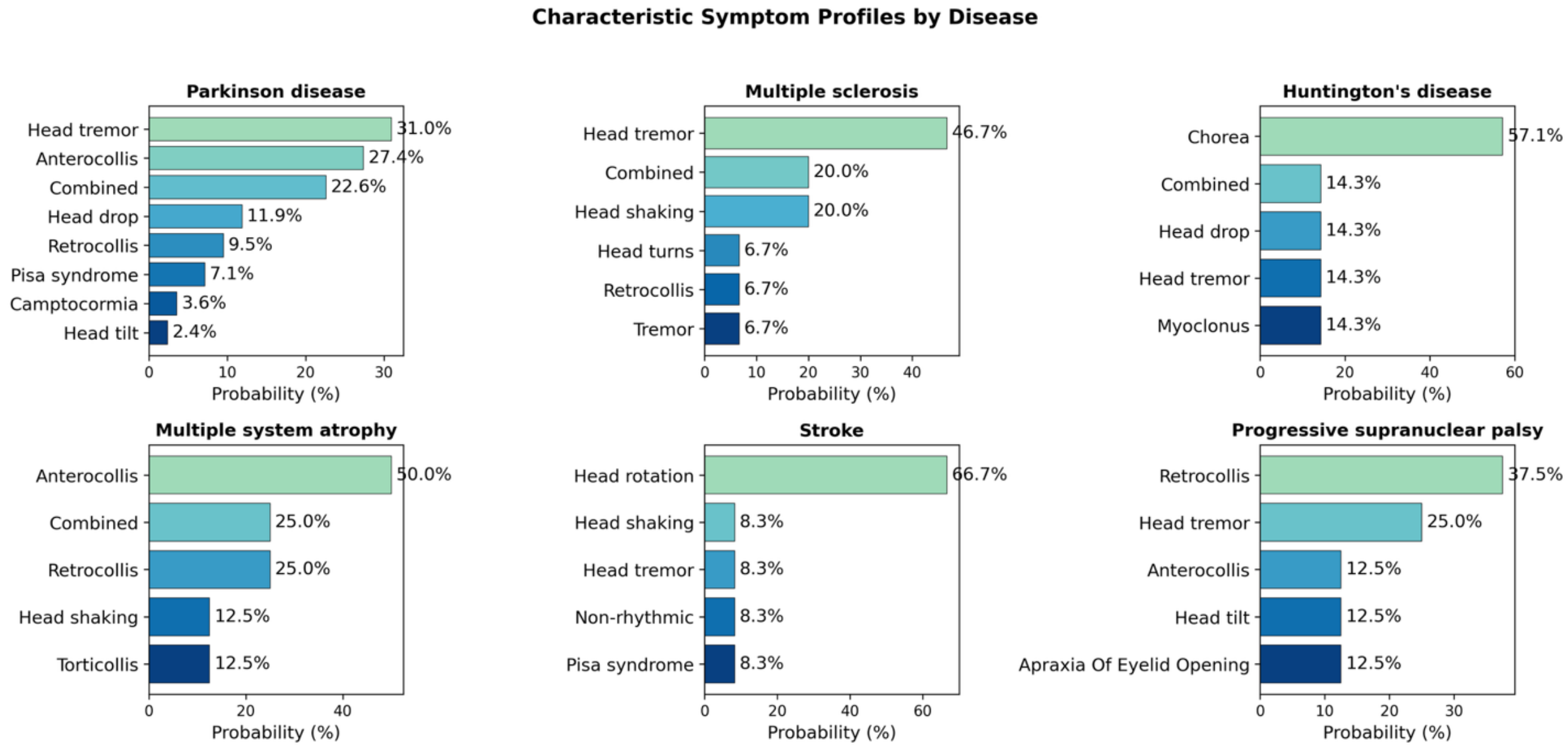

*Figure 15: Characteristic Symptom Profiles by Disease*

### *4.4 Quantitative Movement Features*

Quantitative characterization of AHM through frequency, velocity, amplitude, and latency measurements is essential for differential diagnosis, as these parameters capture key aspects of movement phenomenology. Frequency distinguishes rhythmic pattern [53], amplitude reflects severity and functional impairment, velocity reveals movement dynamics and underlying motor control deficits [54], and latency provides temporal information regarding movement onset and duration critical for understanding pathophysiological mechanisms [55]. Recent advances in AI have enabled integration of these multimodal measurements for automated diagnosis [20].

Table 7 summarizes the prevalence of quantitative movement feature reporting across 230 papers. Frequency and amplitude are the most consistently reported features (53.9% and 54.8% of studies, respectively), reflecting their clinical observability and direct relevance to severity assessment. In contrast, velocity and latency are documented less frequently (31.3% and 33.9%), indicating a gap in capturing movement dynamics and temporal characteristics. These findings highlight the inconsistency in quantitative reporting practices across the AHM literature, with implications for the completeness of knowledge-based datasets such as NeuroPose-AHM.

*Table 7: Distribution of reported quantitative movement features*

| Category | Total Papers | Frequency | Velocity | Amplitude | Latency |
|---|---|---|---|---|---|
| Diseases | 134 | 71 | 53 | 72 | 50 |
| Disorders | 107 | 61 | 23 | 58 | 30 |
| Total | 230 | 124 | 72 | 126 | 78 |

Quantitative feature measurements varied substantially by measurement modality, as shown in Table 8. For cervical dystonia, amplitude alone is reported using three incompatible measurement units: spatial displacement (mm, MRI), angular deviation (°, goniometry), and muscle electrical activity (mV, EMG). Velocity is also reported in both angular (°/s) and linear (m/s) units depending on the tracking method. Similarly, head tremor amplitude in essential tremor is documented using millimeters, centimeters, and power spectral density ($g^2$/Hz). This cross-modality heterogeneity confirms that measurements are not directly comparable without unit-aware normalization, thereby motivating the feature engineering strategy adopted in Task 1.

*Table 8: Method-Stratified Quantitative Movement Features*

| Condition | Movement Type | Method | Velocity | Velocity Unit | Amplitude | Amplitude Unit | Latency | Latency Unit |
|---|---|---|---|---|---|---|---|---|
| Cerebellar ataxia | Head tremor | Accelerometer/IMU; | - | | 75.0 | %Vr | - | |
| | | EEG; EMG | - | | - | | 13.7 | ms |
| Cervical dystonia | Combined | EMG; Goniometer | - | | 45.0-90.0 | degrees | - | |
| | Head tremor | Accelerometer/IMU | 7.8 | °/s | 0.0-0.6 | g²/Hz | 7.0-130.3 | ms |
| | | EMG | - | | 300.0 | µV | 100.0 | ms |
| | | Motion capture | - | | 0.5-0.7 | mm | - | |
| | Torticollis | Accelerometer/IMU; EMG; TMS | - | | 0.9 | mV | 10.1 | ms |
| | | Accelerometer/IMU; TMS | - | | 1.9-2.0 | mV | 36.5-45.8 | ms |
| | | EEG | - | | - | | 500.0 | ms |
| | | EMG | 60.0 | deg/sec | 3.5 | | 103.7 | ms |
| | | EMG; Motion capture | - | | 66.0-66.0 | degrees | - | |
| | | Eye tracking/VOG | 30.0 | ° per second | 10.0 | degrees | - | |
| | | Goniometer | - | | 50.4 | degrees | - | |
| | | MRI | - | | 0.3 | mm | - | |
| | | Motion capture | 29.2-31.6 | °/s | 0.4-45.0 | degrees | - | |
| Essential tremor | Head tremor | Accelerometer/IMU | 12.3 | °/s | 0.1-16.0 | mm | 151.0 | ms |
| | | Eye tracking/VOG; Search coil | 8.0-10.6 | °/s | - | | 155.3-181.5 | ms |
| | | Motion capture | 1.4-1.4 | m/s | 1.3-2.1 | mm | - | |
| | | Video analysis | - | | 0.0-5.0 | cm | - | |
| Multiple sclerosis | Head shaking | Accelerometer/IMU | 62.7-113.6 | °/s | 15.1-44.4 | degrees | - | |
| | | Accelerometer/IMU; Video analysis | 187.5 | °/s | - | | 222.7 | ms |
| | Head tremor | Accelerometer/IMU | - | | 2.0-13.3 | degrees | - | |

| | | | | | | | |
|---|---|---|---|---|---|---|---|
| | | Accelerometer/IMU; EEG; EMG | - | 50.0 | %Vr | - | |
| Parkinson disease | Anterocollis | Goniometer | - | 58.7-62.5 | degrees | - | |
| | Head tremor | Accelerometer/IMU | - | 0.0-0.0 | g | 10.0 | ms |
| | | Accelerometer/IMU; EMG | - | - | | 120.0-150.0 | ms |
| | Pisa syndrome | Goniometer | - | 11.3 | degrees | - | |
| | | Radiography; Video analysis | - | 21.0 | degrees | - | |

*Notes: EEG, electroencephalography; EMG, electromyography; IMU, inertial measurement unit; VOG, video-oculography. Units: °/s, degrees per second; °, degrees; mm, millimeters; cm, centimeters; m/s, meters per second; $g^2$/Hz, power spectral density; mV, millivolts; μV, microvolts; %Vr, percentage of voluntary range of motion; ms, milliseconds.*

### *4.5 Clinical Assessment Scales*

While quantitative movement features provide direct measurements of AHM phenomenology, clinical practice often relies on standardized assessment scales such as TWSTRS and UPDRS, which incorporate both motor subscales and non-motor components related to pain, disability, and quality of life. The distribution of these scales across the NeuroPose-AHPOS corpus reveals important patterns in severity assessment across neurological conditions. Among 396 papers reporting clinical assessment data, 33 unique motor assessment scales relevant to AHM detection were identified. Scale usage was highly concentrated, with the 15 most frequently employed scales accounting for the majority of reported assessments, as shown in Figure 16. The WSTRS was the most frequently reported scale, appearing in 81 studies, followed by the Tsui scale for cervical dystonia and the UPDRS and Hoehn–Yahr scales for Parkinson's disease.

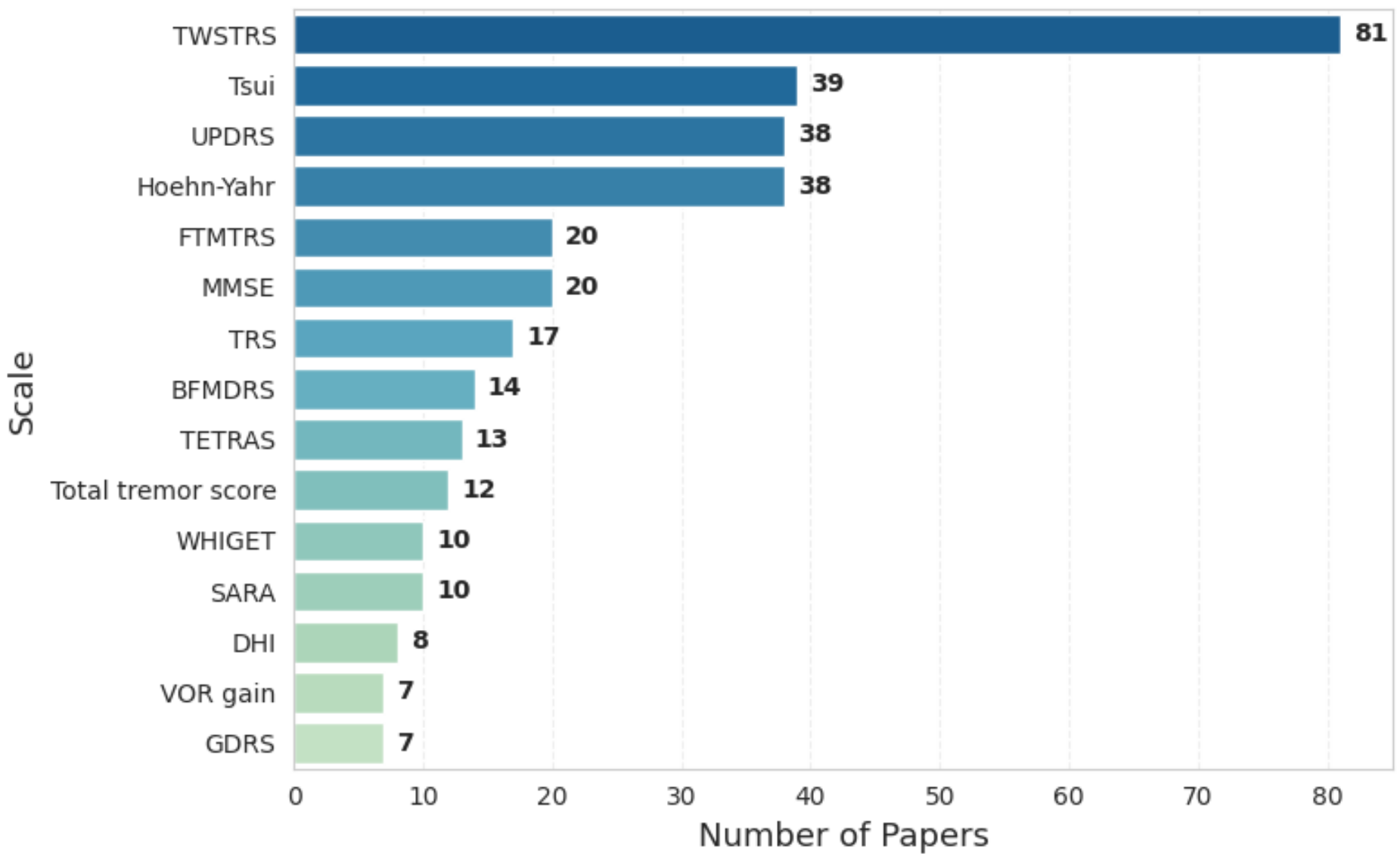

*Figure 16: Frequency distribution of the most mentioned clinical assessment*

Clinical assessment of AHM relies on standardized rating scales to quantify movement severity and functional impact. Table 9 summarizes the primary motor assessment scales commonly reported in neurological disorders. Three patterns are of particular analytical relevance. Cervical dystonia shows the highest degree of standardization, with TWSTRS dominating, followed by the Tsui scale, providing a reliable severity basis for HNSI construction in Task 2. Essential tremor exhibits greater heterogeneity, with multiple tremor-specific scales used at comparable frequencies. Across other conditions, scale selection is largely condition-specific, with Hoehn–Yahr for Parkinson's disease and instruments such as EDSS, SARA, and ICARS used for multiple sclerosis and cerebellar disorders. These patterns reflect disease-specific assessment priorities rather than a unified AHM-oriented severity framework.

*Table 9: Distribution of clinical assessment scales used to evaluate AHM across neurological conditions.*

| Condition | Total Papers | Scale Name | Papers |
|---|---|---|---|
| Cervical dystonia | 129 | TWSTRS | 80 |
| | | Tsui | 37 |
| | | TETRAS | 4 |
| | | TRS | 4 |
| | | SARA | 4 |
| Essential tremor | 66 | TETRAS | 13 |
| | | TRS | 12 |
| | | Total tremor score | 12 |
| | | WHIGET | 10 |
| | | FTM | 5 |
| Parkinson disease | 41 | Hoehn-Yahr | 34 |
| | | C2 slope angle † | 2 |
| | | Cervical sagittal vertical axis † | 2 |
| | | Cervical lordosis angle † | 2 |
| | | QMG † | 2 |
| Multiple sclerosis | 7 | EDSS | 5 |
| | | CV of aVOR gain† | 1 |
| | | Expanded Disability Status Scale | 1 |
| | | MRC | 1 |
| | | aVOR gain† | 1 |
| Cerebellar ataxia | 5 | SARA | 3 |
| | | Hoehn-Yahr | 1 |
| | | ICARS | 1 |
| | | VOR gain† | 2 |

†Postural or functional assessment measures used in the absence of condition-specific AHM severity instruments

### *4.6 Cervical Dystonia: Detailed Characterization*

#### *4.6.1 Task 1: Abnormal Head Movement Classification*

The comparative evaluation of classifier performance reveals clear differences in robustness under class imbalance. As summarized in Table 10, models are assessed using macro-averaged F1 as the primary metric due to its sensitivity to minority-label performance. LR demonstrates the most balanced performance across labels, achieving the highest macro F1, strong recall, and the lowest Hamming loss, indicating robust per-label prediction accuracy and stable probability calibration.

XGBoost shows strong ranking capability, reflected in the highest ROC-AUC and competitive Hamming loss, although its lower macro F1 suggests reduced robustness under threshold-optimized multi-label evaluation. RF achieves moderate performance but shows weaker handling of minority labels. LightGBM exhibits the lowest exact match accuracy and the highest Hamming loss, indicating limited suitability for this dataset. The MLP achieves high precision but comparatively lower recall, reflecting conservative prediction behavior under class imbalance.

*Table 10: Overall Performance of Classifiers on AHM Classification*

| Model | Accuracy | Precision | Recall | F1 | ROC-AUC | Hamming loss |
|---|---|---|---|---|---|---|
| RF | 0.5122 | 0.7159 | 0.8077 | 0.7590 | 0.7706 | 0.1951 |
| XGBoost | 0.5366 | 0.8194 | 0.7564 | 0.7867 | 0.8850 | 0.1561 |
| LightGBM | 0.3659 | 0.7195 | 0.7564 | 0.7375 | 0.8377 | 0.2049 |
| LR | 0.6829 | 0.7789 | 0.9487 | 0.8555 | 0.8556 | 0.1220 |
| MLP | 0.4878 | 0.8095 | 0.6538 | 0.7234 | 0.7478 | 0.1902 |

To further examine model behavior across individual movement types, per-label F1 scores are reported in Table 11 and visualized in Figure 17. Performance on torticollis is consistently high across models, reflecting its dominant representation in the dataset. Head tremor is also classified reliably, likely due to its distinct oscillatory kinematic profile. LR substantially outperforms other models on retrocollis, whereas tree-based and boosting methods (RF, XGBoost, LightGBM) exhibit markedly lower performance, indicating limited sensitivity to this severely imbalanced class. A similar pattern is observed for anterocollis, where LR again achieves the strongest performance, followed by MLP and XGBoost. These results suggest improved minority-label sensitivity under balanced weighting and linear regularization.

*Table 11: Per-Label F1 Scores Across AHM Types*

| Label | RF | XGBoost | LightGBM | LR | MLP |
|---|---|---|---|---|---|
| Torticollis | 0.9333 | 0.9333 | 0.9153 | 0.9333 | 0.8627 |
| Laterocollis | 0.8387 | 0.8387 | 0.8387 | 0.8649 | 0.5926 |
| Anterocollis | 0.5185 | 0.6364 | 0.6087 | 0.7500 | 0.7273 |
| Retrocollis | 0.4286 | 0.2500 | 0.2500 | 0.7500 | 0.5000 |
| Head Tremor | 0.9000 | 0.8571 | 0.7826 | 0.9000 | 0.7200 |

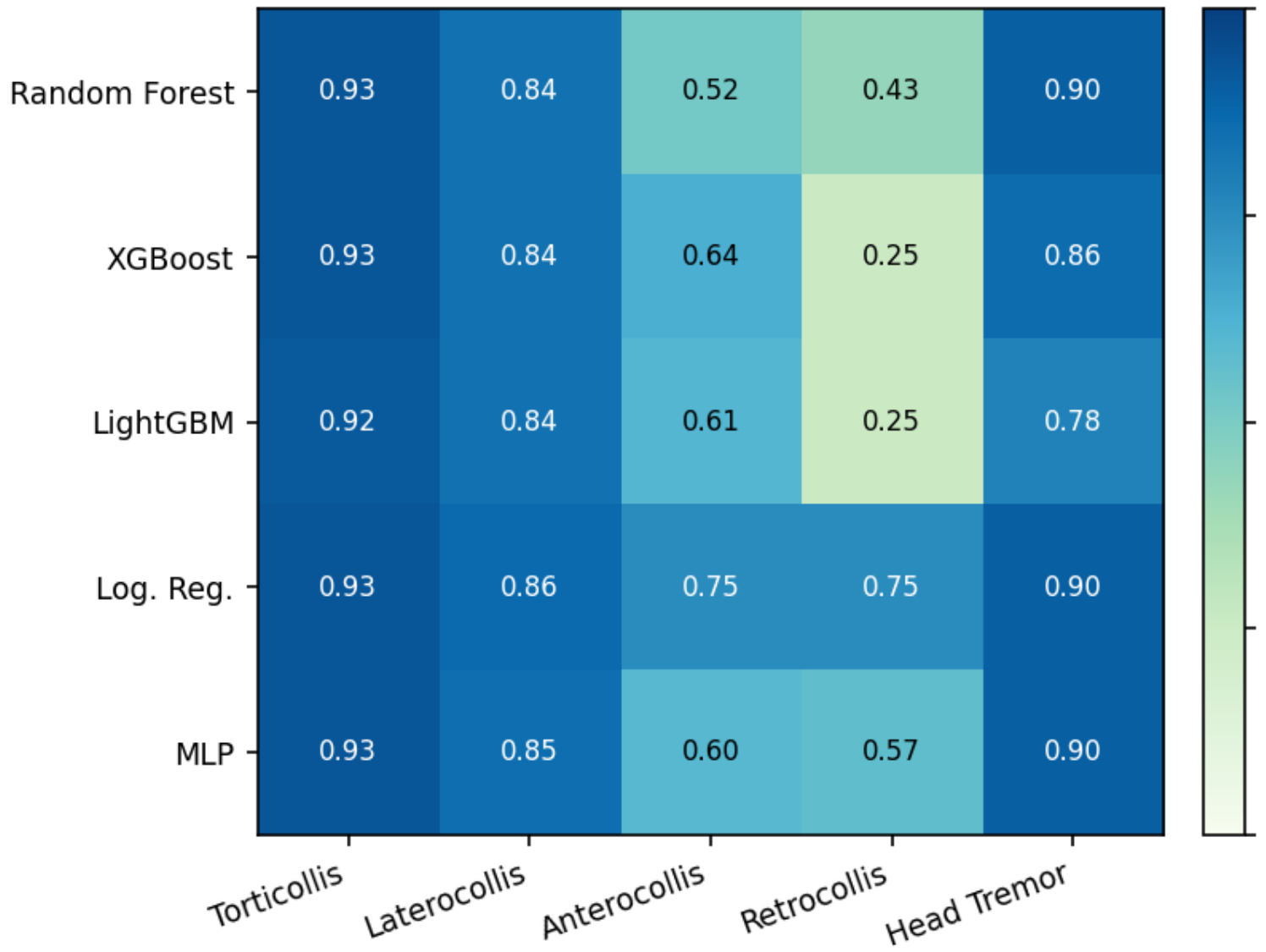

*Figure 17: Per-Label F1 Heatmap Visualization*

### *4.6.2 Task 2: Head-Neck Index Construction*

Scale coverage and the resulting HNSI ranges are summarized in Table 12. TWSTRS, as the most frequently reported scale, yields a broad normalized severity range, whereas the Tsui scale produces a narrower distribution. In contrast, the TRS shows the widest upper-bound severity and includes representation of more severe clinical cases, while the GDRS demonstrates a constrained range consistent with its limited presence in the sampled corpus.

Table 12: HNSI Scale Coverage

| **Scale** | **Papers** | **Records** | **HNSI Range** |
|---|---|---|---|
| TWSTRS | 37 | 161 | 0.043 – 0.771 |
| Tsui | 24 | 72 | 0.063 – 0.460 |
| TRS | 6 | 13 | 0.071 – 0.792 |
| GDRS | 3 | 34 | 0.538 – 0.621 |
| **HNSI** | **66** | **278** | **0.043 – 0.792** |

At the paper level, the HNSI distribution spans a broad severity range across the literature corpus. Classification into predefined severity bands indicates that most studies fall within the mild and moderate categories, with fewer representing severe cases. After aggregation across instruments and application of the eligibility criteria, the resulting HNSI distribution forms the quantitative basis for the clinical validation presented in Task 3 and establishes the baseline severity landscape for benchmarking real-world patient cohorts. Table 13 summarizes the HNSI severity band distribution.

*Table 13: HNSI Severity Band Distribution*

| Severity Band | HNSI Range | Papers (n) | Percentage | Mean HNSI |
|---|---|---|---|---|
| Mild | < 0.33 | 31 | 47% | 0.186 |
| Moderate | 0.33 – 0.66 | 28 | 42% | 0.456 |
| Severe | ⩾ 0.66 | 7 | 11% | 0.747 |

### *4.6.3 Task 3: Clinical Validation Against Real-World Patient Data*

A severity band analysis is conducted to compare the distributions of the external clinical dataset and the HNSI literature index. The external clinical dataset serves as the reference distribution representing independently observed real-world CD severity. The HNSI literature distribution is evaluated against this reference to assess whether the literature-derived index reproduces severity patterns observed in clinical practice. Both sources generate scores on the same 0–1 interval through identical normalization of the TWSTRS severity subscale. Table 14 presents the comparative distribution of severity bands across the two samples.

Table 14: Comparative HNSI Severity Band Distribution

| Severity Band | Threshold | Dataset | HNSI Literature |
|---|---|---|---|
| Mild | HNSI < 0.33 | 3.3% | 29.1% |
| Moderate | 0.33 ⩽ HNSI < 0.66 | 90.0% | 56.5% |
| Severe | HNSI ⩾ 0.66 | 6.7% | 6.7% |

In the external dataset, the severity distribution is predominantly concentrated in the moderate band (90.0%), with one patient (3.3%) classified within the mild severity band and two patients (6.7%) within the severe band. Raw TWSTRS scores range from 8 to 28 out of 35, with a mean normalized severity of 0.538 (SD = 0.112). This distribution is consistent with the clinical profile of a treatment-engaged population. As documented in the source study, all 30 patients were actively receiving botulinum toxin therapy at the time of assessment, indicating that mild cases are structurally underrepresented due to clinical selection rather than chance.

The severe severity band (HNSI ⩾ 0.66; corresponding to TWSTRS ⩾ 23.1) provides the most informative comparison, as severe cases are present in both sources and are minimally affected by the selection bias described above. Within the external cohort, 6.7% of patients are classified as severe, while the HNSI literature distribution yields the same proportion. This finding is therefore interpreted as a preliminary plausibility indication that the literature-derived index can be appropriately at the severe end of the severity spectrum, subject to confirmation in larger cohorts.

Two methodological properties support the validity of this comparison. First, the data sources are independent: the external dataset was collected at a single hospital in Japan using motion capture–based severity assessment, whereas the HNSI was derived through systematic extraction from peer-reviewed studies across multiple countries, clinical environments, and time periods. Second, the comparison is construct-aligned, as both sources rely on the TWSTRS severity subscale normalized by the same theoretical maximum. The comparable severe-band proportions provide a descriptive indication that the HNSI may be calibrated at the severe end of the CD severity spectrum.

#### *4.6.4 Task 4: Bridge Analysis and Cross-Component Validation*

The correlation analysis between HNSI scores and movement-type probability outputs across the overlapping subset of papers is summarized in Table 15. It presents positive correlations between HNSI and movement-type probabilities across models, which are consistent in both direction and magnitude across all three classifiers. Given the exploratory nature of this analysis (n = 24), 95% confidence intervals, computed using Fisher's z-transformation [56], are reported alongside each coefficient, and the findings are interpreted as preliminary indicators of internal coherence within the analytical framework.

Table 15: Correlation Analysis Between HNSI Scores and Movement-Type Probabilities

| Model | Feature | r | p-value | Significant | 95% CI |
|---|---|---|---|---|---|
| **LGBM** | Mean probability | 0.578 | 0.003 | $p<0.01$ | [0.23, 0.80] |
| | Laterocollis prob | 0.530 | 0.008 | $p<0.01$ | [0.16, 0.77] |
| | Retrocollis prob | 0.450 | 0.027 | $p<0.05$ | [0.06, 0.72] |
| **LR** | Mean probability | 0.532 | 0.007 | $p<0.01$ | [0.16, 0.77] |
| | Anterocollis prob | 0.595 | 0.002 | $p<0.01$ | [0.25, 0.81] |
| | Laterocollis prob | 0.457 | 0.025 | $p<0.05$ | [0.07, 0.73] |
| **MLP** | Mean probability | 0.760 | <0.001 | $p<0.001$ | [0.51, 0.89] |
| | Laterocollis prob | 0.779 | <0.001 | $p<0.001$ | [0.55, 0.90] |
| | Anterocollis prob | 0.740 | <0.001 | $p<0.001$ | [0.48, 0.88] |

The MLP probability outputs produce the strongest bridge signal among the three evaluated models in this exploratory cross-task analysis. Continuous probability estimates show the strongest associations with HNSI scores, particularly for laterocollis, anterocollis, and the mean probability across labels. This pattern is consistent with the functional properties of continuous probability outputs. For example, a paper in which laterocollis is highly prevalent across patient groups produces a higher predicted probability than a paper in which the pattern is infrequent. This preserved probabilistic gradient appears to align closely with the independently derived HNSI severity scores.

Stronger bridge correlations for the MLP do not imply superior classification performance. The enhanced correlation strength reflects the model's diffuse and continuous probability distributions rather than diagnostic accuracy in label assignment. For clinical classification purposes, LR remains the recommended model. The MLP bridge findings instead indicate that its probability outputs capture severity-related variance within the kinematic feature space, suggesting potential utility for severity-graded ranking or prioritization tasks. These represent complementary, rather than competing clinical functions.

LR emerges as the second strongest bridge model where positive associations are observed for anterocollis probability and mean probability. This convergence is analytically significant. LR is both the highest-performing classifier overall and the second strongest model in the bridge analysis. The linear feature weighting that supports its classification performance also preserves severity-related structure within the kinematic feature space. The alignment between classification performance and independent severity correlation provides construct-level coherence, indicating that the kinematic feature space captures severity-related structure independently of the classification objective.

## 5. Conclusion

Neurologically induced abnormal head movements (AHMs) remain a fragmented research domain, with clinical evidence dispersed across heterogeneous studies that employ inconsistent terminology, measurement standards, and reporting conventions. This study introduces NeuroPose-AHM, a knowledge-based dataset constructed from peer-reviewed neurological literature through a multi-LLM extraction framework. The dataset integrates quantitative kinematic measurements, clinical severity scores, and patient-group-level characteristics across multiple neurological conditions, enabling more computational analysis of AHMs.

To demonstrate its analytical utility, a four-task framework was applied to cervical dystonia, the neurological condition most directly associated with pathological head posture. The results show that literature-derived kinematic features support reliable multi-label classification of AHM types, while heterogeneous clinical rating scales can be unified through the proposed Head–Neck Severity Index (HNSI). Preliminary comparison against an independent clinical dataset provides a plausibility indication that the HNSI may be calibrated at the severe end of the severity spectrum, and bridge analysis further supports internal consistency between movement classification outputs and severity estimates.

Several limitations constrain the current findings. Class imbalance in the CD-Quantitative subset constrained classification performance for minority movement types. The clinical validation and bridge analyzes are further limited by the small sample sizes, including a cohort of 30 patients from a single clinical site and an overlapping subset of 24 papers, which restricts the generalizability of the findings.

Future work will address these limitations through several directions. Expanding the NeuroPose-AHM dataset with additional studies and conditions will help reduce class imbalance and improve model sensitivity for minority movement types. Larger multi-center clinical cohorts will be incorporated to further validate and benchmark the framework on a broader population of cervical dystonia patients, improving the generalizability of severity assessment. Moreover, NeuroPose-AHM is intended to serve as an evolving foundation that brings AI-driven diagnosis of neurological AHMs closer to clinical application.

## References

[1] Birdwhistell RL. Introduction to kinesics: an annotation system for analysis of body motion and gesture. 1952.

[2] Akbari MR, Khorrami-Nejad M, Kangari H, Heirani M, Baghban AA, Raeesdana K, et al. Does head tilt influence facial appearance more than head turn? J Ophthalmic Vis Res 2023;18:297.

[3] Akbari MR, Khorrami-Nejad M, Kangari H, Akbarzadeh Baghban AA, Ranjbar Pazouki MR. Ocular abnormal head posture: A literature review. J Curr Ophthalmol 2021;33:379–87. https://doi.org/10.4103/joco.joco_114_20.
[4] Prudente CN, Hess EJ, Jinnah HA. Dystonia as a network disorder: what is the role of the cerebellum? Neuroscience 2014;260:23–35.
[5] Albanese A, Bhatia K, Bressman SB, others. Phenomenology and classification of dystonia: a consensus update. Movement Disorders 2013;28:863–73. https://doi.org/10.1002/mds.25475.
[6] Thayani M, Jinnah HA. Can symptoms or signs of cervical dystonia occur without abnormal movements of the head or neck? Parkinsonism Relat Disord 2024;123:106958.
[7] Velickovic M, Benabou R, Brin MF. Cervical dystonia: pathophysiology and treatment options. Drugs 2001;61:1921–43.
[8] Zhang Z, Cisneros E, Lee H, Vu JP, Chen QY, Benadof CN, et al. Hold that pose: capturing cervical dystonia's head deviation severity from video. Ann Clin Transl Neurol 2022;9:684–94. https://doi.org/10.1002/acn3.51549.
[9] Neychev VK, Fan X, Mitev VI, Hess EJ, Jinnah HA. The basal ganglia and cerebellum interact in the expression of dystonic movement. Brain 2008;131:2499–509.
[10] Loens S, Verrel J, Herrmann V-M, Kienzle A, Tzvi E, Weissbach A, et al. Motor learning deficits in cervical dystonia point to defective basal ganglia circuitry. Sci Rep 2021;11:7332.
[11] Udaka YT, Packer RJ. Pediatric brain tumors. Neurol Clin 2018;36:533–56.
[12] Renne B, Rueckriegel S, Ramachandran S, Radic J, Steinbok P, Singhal A. Bobble-head doll syndrome: report of 2 cases and a review of the literature, with video documentation of the clinical phenomenon. J Neurosurg Pediatr 2018;21:236–46.
[13] Qureshi FH, Qureshi SH, Zia T, Khawaja F. Huntington's disease (HD): a brief review. European Journal of Public Health Studies 2022;5.
[14] Defazio G, Jankovic J, Giel JL, Papapetropoulos S. Descriptive epidemiology of cervical dystonia. Tremor and Other Hyperkinetic Movements 2013;3:tre-03.
[15] Tiderington E, Goodman EM, Rosen AR, Hapner ER, Johns III MM, Evatt ML, et al. How long does it take to diagnose cervical dystonia? J Neurol Sci 2013;335:72–4.
[16] Albanese A, Bhatia KP, Cardoso F, Comella C, Defazio G, Fung VSC, et al. Isolated Cervical Dystonia: Diagnosis and Classification. MOVEMENT DISORDERS 2023;38:1367–78. https://doi.org/10.1002/mds.29387.
[17] Song P, Zhang Y, Zha M, Yang Q, Ye X, Yi Q, et al. The global prevalence of essential tremor, with emphasis on age and sex: A meta-analysis. J Glob Health 2021;11:04028.
[18] Jain S, Lo SE, Louis ED. Common misdiagnosis of a common neurological disorder: how are we misdiagnosing essential tremor? Arch Neurol 2006;63:1100–4.
[19] Louis ED. The association between essential tremor and Parkinson's disease: a systematic review of clinical and epidemiological studies. J Clin Med 2025;14:2637.
[20] Vizcarra JA, Yarlagadda S, Xie K, Ellis CA, Spindler M, Hammer LH. Artificial intelligence in the diagnosis and quantitative phenotyping of hyperkinetic movement disorders: a systematic review. J Clin Med 2024;13:7009.
[21] Martínez-García-Peña R, Koens LH, Azzopardi G, Tijssen MAJ. Video-Based Data-Driven Models for Diagnosing Movement Disorders: Review and Future Directions. Movement Disorders 2025;40:2046–66. https://doi.org/10.1002/mds.30327.
[22] Valeriani D, Simonyan K. A microstructural neural network biomarker for dystonia diagnosis identified by a DystoniaNet deep learning platform. Proceedings of the National Academy of Sciences 2020;117:26398–405.

[23] Nakamura T, Sekimoto S, Oyama G, Shimo Y, Hattori N, Kajimoto H. Pilot feasibility study of a semi-automated three-dimensional scoring system for cervical dystonia. PLoS One 2019;14:e0219758.
[24] Ye C, Xiao Y, Li R, Gu H, Wang X, Lu T, et al. Pilot feasibility study of a multi-view vision based scoring method for cervical dystonia. Sensors 2022;22:4642.
[25] Zhang Z, Cisneros E, Lee HY, Vu JP, Chen Q, Benadof CN, et al. Hold that pose: capturing cervical dystonia's head deviation severity from video. Ann Clin Transl Neurol 2022;9:684–94.
[26] Vu JP, Cisneros E, Lee HY, Le L, Chen Q, Guo XA, et al. Head tremor in cervical dystonia: Quantifying severity with computer vision. J Neurol Sci 2022;434. https://doi.org/10.1016/j.jns.2022.120154.
[27] Peach R, Friedrich M, Fronemann L, Muthuraman M, Schreglmann SR, Zeller D, et al. Head movement dynamics in dystonia: a multi-centre retrospective study using visual perceptive deep learning. NPJ Digit Med 2024;7:160.
[28] Pecoraro PM, Marsili L, Espay AJ, Bologna M, di Biase L. Computer vision technologies in movement disorders: A systematic review. Mov Disord Clin Pract 2025;12:1229–43.
[29] Pearce JMS. Disease, diagnosis or syndrome? Pract Neurol 2011;11:91–7.
[30] Thurm A, Srivastava S. On Terms: What's in a name? Intellectual disability and "condition,""disorder,""syndrome,""disease," and "disability." Am J Intellect Dev Disabil 2022;127:349–54.
[31] Introducing Claude Sonnet 4.5 \ Anthropic n.d. https://www.anthropic.com/news/claude-sonnet-4-5 (accessed March 5, 2026).
[32] Singh A, Fry A, Perelman A, Tart A, Ganesh A, El-Kishky A, et al. Openai gpt-5 system card. ArXiv Preprint ArXiv:260103267 2025.
[33] Introducing GPT-5 | OpenAI n.d. https://openai.com/index/introducing-gpt-5/?utm_source=chatgpt.com (accessed March 28, 2026).
[34] Zheng L, Chiang W-L, Sheng Y, Zhuang S, Wu Z, Zhuang Y, et al. Judging llm-as-a-judge with mt-bench and chatbot arena. Adv Neural Inf Process Syst 2023;36:46595–623.
[35] Wang P, Li L, Chen L, Cai Z, Zhu D, Lin B, et al. Large language models are not fair evaluators. Proceedings of the 62nd Annual Meeting of the Association for Computational Linguistics (Volume 1: Long Papers), 2024, p. 9440–50.
[36] Chan C-M, Chen W, Su Y, Yu J, Xue W, Zhang S, et al. Chateval: Towards better llm-based evaluators through multi-agent debate. ArXiv Preprint ArXiv:230807201 2023.
[37] Zeng Z, Yu J, Gao T, Meng Y, Goyal T, Chen D. Evaluating large language models at evaluating instruction following. ArXiv Preprint ArXiv:231007641 2023.
[38] Cohen J. A coefficient of agreement for nominal scales. Educ Psychol Meas 1960;20:37–46.
[39] Shrout PE, Fleiss JL. Intraclass correlations: uses in assessing rater reliability. Psychol Bull 1979;86:420.
[40] Koo TK, Li MY. A guideline of selecting and reporting intraclass correlation coefficients for reliability research. J Chiropr Med 2016;15:155–63.
[41] Jankovic J, Leder S, Warner D, Schwartz K. Cervical dystonia: clinical findings and associated movement disorders. Neurology 1991;41:1088.
[42] Nijmeijer SWR, Koelman J, Kamphuis DJ, Tijssen MAJ. Muscle selection for treatment of cervical dystonia with botulinum toxin–a systematic review. Parkinsonism Relat Disord 2012;18:731–6.
[43] Jost WH, Tatu L. Selection of muscles for botulinum toxin injections in cervical dystonia. Mov Disord Clin Pract 2015;2:224.
[44] Morgan JC, Sethi KD. Drug-induced tremors. Lancet Neurol 2005;4:866–76.

[45] Hess CW, Pullman SL. Tremor: clinical phenomenology and assessment techniques. Tremor and Other Hyperkinetic Movements 2012;2.
[46] Brigato L, Iocchi L. A close look at deep learning with small data. 2020 25th international conference on pattern recognition (ICPR), IEEE; 2021, p. 2490–7.
[47] Shwartz-Ziv R, Armon A. Tabular data: Deep learning is not all you need. Information Fusion 2022;81:84–90.
[48] Cox DR. The regression analysis of binary sequences. J R Stat Soc Series B Stat Methodol 1958;20:215–32.
[49] Breiman L. Random forests. Mach Learn 2001;45:5–32.
[50] Chen T, Guestrin C. Xgboost: A scalable tree boosting system. Proceedings of the 22nd acm sigkdd international conference on knowledge discovery and data mining, 2016, p. 785–94.
[51] Ke G, Meng Q, Finley T, Wang T, Chen W, Ma W, et al. Lightgbm: A highly efficient gradient boosting decision tree. Adv Neural Inf Process Syst 2017;30.
[52] Rumelhart DE, Hinton GE, Williams RJ. Learning representations by back-propagating errors. Nature 1986;323:533–6.
[53] Bhatia KP, Bain P, Bajaj N, Elble RJ, Hallett M, Louis ED, et al. Consensus Statement on the classification of tremors. from the task force on tremor of the International Parkinson and Movement Disorder Society. Movement Disorders 2018;33:75–87.
[54] Bologna M, Paparella G, Fasano A, Hallett M, Berardelli A. Evolving concepts on bradykinesia. Brain 2020;143:727–50.
[55] Panyakaew P, Cho HJ, Lee SW, Wu T, Hallett M. The pathophysiology of dystonic tremors and comparison with essential tremor. Journal of Neuroscience 2020;40:9317–26.
[56] Cohen J, Cohen P, West SG, Aiken LS. Applied multiple regression/correlation analysis for the behavioral sciences. Routledge; 2013.